\documentclass{article}

\usepackage{amsmath}
\usepackage{amssymb}
\usepackage{booktabs}
\usepackage{multirow}
\usepackage{graphicx}
\usepackage{enumitem}
\usepackage{subcaption}
\usepackage{xcolor}
\usepackage{pifont}
\usepackage{makecell}
\usepackage[table]{xcolor}
\usepackage{graphicx}
\usepackage{makecell}
\usepackage{subcaption}
\usepackage{array}

\newcolumntype{P}[1]{>{\centering\arraybackslash}p{#1}}
\newcommand{\obj}[2]{\makecell[c]{\textbf{#1}\\\textbf{#2}}}

\newcommand{\cmark}{\textcolor{green!60!black}{\ding{51}}}
\newcommand{\xmark}{\textcolor{red}{\ding{55}}}

\usepackage[final]{corl_2026} 

\title{$\mathcal{T}(R,O)$ Grasp++: Cross-Embodiment Dexterous Grasping in Cluttered Environments at Scale}

\author{
  \textbf{
    Xin Fei\textsuperscript{1,2},
    Zhixuan Xu\textsuperscript{1,2},
    Huaicong Fang\textsuperscript{3},
    Fangjun Chen\textsuperscript{2,4}
  }
  \\
  \textbf{
    Zhaoyu Mu\textsuperscript{1},
    Rui Yang\textsuperscript{1,2},
    Zhexi Luo\textsuperscript{2,5},
    Lin Shao\textsuperscript{1,2,*}
  }
  \\[6pt]
     {\footnotesize
    \textsuperscript{1}Department of Computer Science, National University of Singapore
    \quad
    \textsuperscript{2}RoboScience}
    \\
    {\footnotesize
    \textsuperscript{3}College of Control Science and Engineering, Zhejiang University, China}
    \\
    {\footnotesize
    \textsuperscript{4}Department of Electronic and Electrical Engineering, Southern University of Science and Technology, China}
    \\
    {\footnotesize
    \textsuperscript{5}School of Computer Science and Engineering, Sun Yat-sen University, China}
    \\[2pt]
    {\footnotesize
    \textsuperscript{*}Corresponding author} \\
    {\footnotesize
    \href{https://nus-lins-lab.github.io/TROGrasp2Web/}
    {\textcolor{red}{\nolinkurl{https://nus-lins-lab.github.io/TROGrasp2Web/}}}
}
}
\begin{document}
\maketitle
\vspace{-20pt}
\begin{abstract}
Dexterous grasping in cluttered scenes remains a central challenge in robotic manipulation, as it requires high-dimensional hand coordination under complex object-environment constraints. To address this problem, we introduce one of the largest simulation benchmarks to date, comprising around 0.5B grasp annotations across eight dexterous hands and approximately 50K objects. Furthermore, we propose $\mathcal{T}(R,O)$ Grasp++, a scalable diffusion-based framework for cross-embodiment dexterous grasp synthesis in cluttered environments. Extensive experiments demonstrate that our model significantly outperforms existing baselines, achieving an average success rate of 88\% across eight dexterous embodiments. Meanwhile, our pipeline achieves only 0.3s inference latency while improving inference sample efficiency by over 10 times.
\end{abstract}

\vspace{-10pt}

\begin{figure}[htbp]
      \centering
      \includegraphics[width=\linewidth]{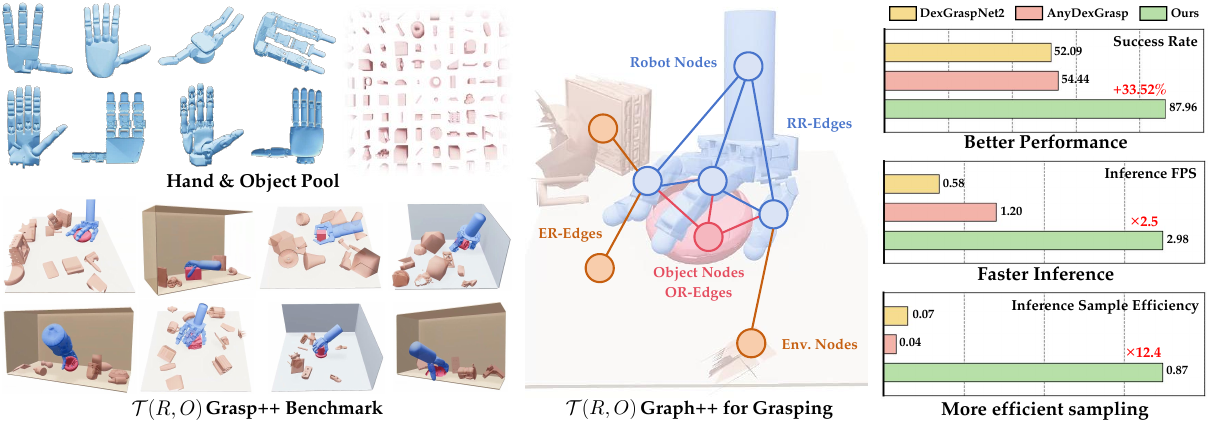}
      \vspace{-15pt}
      \caption{\small Our benchmark comprises around 0.5 billion grasp annotations in cluttered scenes across diverse objects and dexterous hands. $\mathcal{T}(R,O)$ Grasp++ models object-environment-hand spatial transformations with $\mathcal{T}(R,O)$ Graph++, achieving better performance, faster inference, and more efficient sampling.}
      \label{fig:teaser}
  \end{figure}
\vspace{-12pt}
\section{Introduction}
\label{sec:intro}
\vspace{-8pt}
Dexterous grasping in cluttered environments is essential for practical robotic manipulation. However, it remains challenging due to the high-dimensional action space and complex interactions among object, hand and surrounding environments. Existing methods typically generate multiple grasp proposals directly in the scene and rely on collision checking or grasp scoring to select executable grasps~\cite{zhang2024dexgraspnet, zhang2025contactdexnet, hgcnet, fang2025anydexgrasp, DDGC, multimodal2021, zhang2025cadgrasp}. However, previous works have three key limitations. First, the proposal-and-selection paradigm suffers from low inference sample efficiency. Since the relations among the hand, object, and surrounding environment are not sufficiently encoded, many generated proposals are either in collision or dynamically unstable, requiring a large number of candidates to obtain a single successful grasp. Second, model performance is limited by insufficient data scale and diversity. Existing datasets are limited in object and embodiment diversity, with redundant annotations concentrated on a small set of objects~\cite{zhang2024dexgraspnet, zhang2025contactdexnet}. Third, existing grasp representations are insufficient. Prior methods either rely on hand-specific representations, such as hand joint values~\cite{zhang2024dexgraspnet, hgcnet} or predefined grasp templates~\cite{fang2025anydexgrasp, zhang2025cadgrasp}, which limit generalization across hands, or adopt object-centric representations for grasp optimization~\cite{zhang2025contactdexnet, jiang2021graspTTA}, which is computationally expensive.

To address these limitations, we first construct a large-scale dataset across multiple embodiments for dexterous grasping in cluttered environments. Our benchmark contains around 0.5B grasp annotations in common manipulation scenes, with hierarchical difficulty scores that enable fine-grained evaluation across varying grasp difficulties. In addition, we propose $\mathcal{T}(R,O)$ Graph++, a unified grasp representation that models spatial transformations among environment, robot, and object. Built upon $\mathcal{T}(R,O)$ Graph++, our $\mathcal{T}(R,O)$ Grasp++ model performs unified diffusion-based cross-embodiment grasping synthesis. In extensive experiments, our model achieves an average success rate of 87.96\% across eight dexterous hands, an inference speed of 0.34 seconds on an NVIDIA 4090 GPU, and over 10$\times$ higher inference sample efficiency than existing baselines. 

In conclusion, our contributions are summarized as follows: 
(1) We construct a large-scale cluttered-scene dexterous grasping benchmark with 53,675 objects, around 0.5B grasp annotations, and difficulty scores for evaluation under varying grasp difficulties;
(2) We propose $\mathcal{T}(R,O)$ Grasp++, a scalable diffusion-based framework that integrates object-environment geometry with hand kinematics for cross-embodiment grasp synthesis in cluttered environments;
(3) We conduct extensive simulation and real-world experiments, where $\mathcal{T}(R,O)$ Grasp++ significantly outperforms all baselines by over 30\%, while achieving fast and sample-efficient inference;
(4) We present the first scaling study for cross-embodiment dexterous grasping, leveraging our large-scale benchmark to reveal how performance scales with training object diversity and model capacity.
\vspace{-5pt}
\section{Related Works}
\label{sec:rw}

\vspace{-7pt}
\subsection{Dexterous Grasping Dataset}
\label{sec:2.1}
\vspace{-7pt}

Creating large-scale datasets for dexterous grasping is challenging due to the high-dimensional hand action space and complex hand-object contacts. One line of work constructs such datasets through optimization-based synthesis, where grasp poses are directly optimized with force-closure, contact, penetration, and stability objectives~\cite{zhang2024dexgraspnet,zhang2025contactdexnet, wang2022dexgraspnet, pmlr-v205-wu23b}. Another line of work trains grasping policies in simulation and rolls out on large object sets, collecting successful executions as grasp annotations~\cite{zhang2024graspxl,casas2024multigrippergrasp, pmlr-v205-chen23b}. In our work, we train reinforcement learning policies for each dexterous hand and use successful rollouts to annotate 444 million grasps across large-scale object datasets. As shown in Tab.~\ref{tab:dataset}, our dataset covers 53,675 objects and 8 dexterous hands, while additionally providing per-grasp difficulty scores and a mount link to approximate real-world hand-arm deployment.

\begin{table*}[t]
    \centering
    \scriptsize
    \renewcommand\arraystretch{1.2}
    \setlength{\tabcolsep}{4pt}
    \resizebox{\textwidth}{!}{
        \begin{tabular}{c|ccccccc}
            \toprule
            \textbf{Dataset}
            & Obj. Num.
            & Hand Num.
            & Grasp Num.
            & Grasps per Obj.-Hand
            & Cluttered Scenes
            & Hierarchical Annotations
            & Hand Mount
            \\
            \midrule
            GraspXL~\cite{zhang2024graspxl}
                & $>$500 K & 4 & $>$10 M & N.A. & \xmark & \xmark & \xmark
            \\
            Dex1B~\cite{ye2025dex1b}
                & 6 K & 3 & 1 B & N.A. & \xmark & \xmark & \xmark
            \\
            MultiGripperGrasp~\cite{casas2024multigrippergrasp}
                & 345 & 11 & 30.4 M & $\sim$ 8K & \xmark & \xmark & \xmark
            \\
            ContactDexNet~\cite{zhang2025contactdexnet}
                & $>$1.5 K & 1 & N.A. & N.A. & \cmark & \xmark & \xmark
            \\
            DexGraspNet2~\cite{zhang2024dexgraspnet}
                & 1,319 & 1 & 427 M & $\sim$ 324K & \cmark & \xmark & \xmark
            \\
            \textbf{Ours}
                & 53,675 & 8 & 444 M & $\sim$ 1K & \cmark & \cmark & \cmark
            \\
            \bottomrule
        \end{tabular}
    }
    \vspace{-5pt}
    \caption{\small \textbf{Comparison of Dexterous Grasping Dataset.}}
    \label{tab:dataset}
    \vspace{-18pt}
\end{table*}

\vspace{-5pt}
\subsection{Learning-based Dexterous Grasping Synthesis}
\label{sec:2.2}
\vspace{-7pt}

Learning-based frameworks for dexterous grasping synthesis can be categorized into single-object and scene-aware approaches. Single-object methods typically predict or optimize grasp poses for an isolated target object without modeling environment constraints~\cite{zhang2025contactdexnet, wei2024dro, fei2025t, wan2023unidexgrasp++, shao2020unigrasp, xu2024manifoundation, dfc, li2023gendexgrasp, attarian2023geometry, khargonkar2023neuralgrasps}. While these methods achieve promising results in single-object settings, their inference sample efficiency drops substantially when applied to real-world cluttered environments, as many predicted grasps suffer from severe collisions with surrounding objects or environment.

In comparison, scene-aware methods further incorporate environment constraints into grasping synthesis~\cite{zhang2024dexgraspnet, zhang2025contactdexnet, hgcnet, fang2025anydexgrasp, pmlr-v205-chen23b, chen2025clutterdexgrasp, pmlr-v270-wang25l, pmlr-v155-breyer21a}. Despite their progress, existing methods still face limitations in both task setting and grasp representation. In terms of task setting, existing methods generate grasp proposals over the entire scene and then rely on collision checking, grasp scoring, or feasibility filtering to select executable grasps~\cite{zhang2024dexgraspnet, zhang2025contactdexnet, hgcnet, fang2025anydexgrasp, DDGC, multimodal2021}. While suitable for open-world scene grasping, this setting suffers from poor sample efficiency in target-object grasping, as many generated proposals correspond to irrelevant objects or collide with the surrounding environment. As for grasp representation, prior methods either directly predict hand-specific joint values~\cite{zhang2024dexgraspnet,fang2025anydexgrasp, chen2025clutterdexgrasp}, which limits cross-embodiment generalization, or rely on object-centric predictions that require additional slow optimization to recover executable joint configurations~\cite{zhang2025contactdexnet, zhang2025cadgrasp}. To overcome these limitations, we propose $\mathcal{T}(R,O)$ Graph++ as a unified grasp representation across embodiments, enabling direct grasp prediction on the target object in cluttered environments within a unified framework.
\vspace{-5pt}
\section{Benchmark}
\label{sec:benchmark}
\vspace{-5pt}

We present a large-scale benchmark for cross-embodiment dexterous grasping in diverse cluttered scenes, with grasp difficulty assessed by object-level graspability and scene-level clutter complexity to support fine-grained evaluation across varying grasp difficulties. Our dataset comprises 53,675 objects from UniDexGrasp~\cite{xu2023unidexgrasp} and ObjaverseXL~\cite{deitke2023objaverse}, including diverse cluttered scenes in tabletop, shelf, and box settings (as shown in Fig.~\ref{fig:teaser}). Across eight dexterous hands, we annotate around 0.5B valid grasp samples in simulation.

\subsection{Object Collection and Scene Synthesis}
\label{sec:3.1}
\vspace{-5pt}
We collect 13,049 objects from the UnidexGrasp dataset~\cite{xu2023unidexgrasp} and 40,626 objects from the ObjaverseXL dataset~\cite{deitke2023objaverse}. For training, we use dexterous grasp samples in cluttered scenes across all dexterous hands on 47,737 objects. For testing, we use the remaining 5,938 unseen objects to construct cluttered environments for each dexterous hand. In each scene, 15 to 25 objects are randomly placed on a supporting plane. We construct three common manipulation environments: tabletop, shelf and box, with environment scale randomly sampled within predefined ranges.

\subsection{Dexterous Grasp Annotation}
\label{sec:3.2}
\vspace{-5pt}
To efficiently annotate dexterous grasps across robotics hands, we adopt a three-stage annotation pipeline. First, we train embodiment-specific reinforcement learning policies to roll out power and precision grasps on floating objects. Second, we generate stable object poses by randomly sampling initial object poses and dropping the objects onto a tabletop in simulation. Each floating-object grasp is then transformed according to the resulting object pose, and candidates with tabletop penetration are filtered out. Finally, we construct diverse cluttered manipulation scenes across tabletop, shelf, and box settings by placing additional obstacle objects outside the 2D convex hull of the hand-object grasp configuration to avoid collisions. In this way, we annotate approximately 1K grasps in clutter per target object for each dexterous hand.

To better reflect real-world deployment constraints, we attach a cylinder mount link to each palm to approximate the connecting arm segment and include it in collision checking, ensuring sufficient clearance for both the hand and its mounting structure. Our RL-based annotation pipeline supports scalable data generation through massively parallel simulation. On an NVIDIA RTX 4090 GPU, a trained policy generates 216 grasp annotations per second while also achieving a higher filtering success rate compared with optimization-based methods~\cite{li2023gendexgrasp, chen2024bodex}.

\subsection{Hierarchical Difficulty Annotation}
\label{sec:3.3}
\vspace{-5pt}
Beyond grasp annotations, we further assess the execution difficulty of each grasp to support fine-grained evaluation across varying grasp difficulties. Specifically, object-level difficulty $D_{\text{obj}}$ captures the intrinsic graspability of the target object, estimated by the number of RL optimization steps required for reliable grasping. Environment-level difficulty $D_{\text{env}}$ measures local clutter around the target object based on obstacle proximity and spatial enclosure. We combine these two factors as an overall difficulty score for each grasp annotation. Detailed definitions are provided in the Appendix.

\subsection{Dexterous Grasp Evaluation in Simulation}
\label{sec:3.4}
\vspace{-5pt}
We evaluate dexterous grasps in Isaac Gym simulation. A grasp is considered successful if it is collision-free with the surrounding environment and lifts the target object above a predefined height threshold. The hand is initialized at the outer grasp pose, driven to the inner grasp pose with position control, and then lifted upward. For consistency, we use the same simulation protocol for both training data filtering and model evaluation across our method and all baselines.

\section{\texorpdfstring{$\mathcal{T}(R,O)$ Grasp++}{T(R,O) Grasp++} for Dexterous Grasping in Clutter}
\label{sec:model}

\begin{figure}[t]
    \centering
    \includegraphics[width=\linewidth]{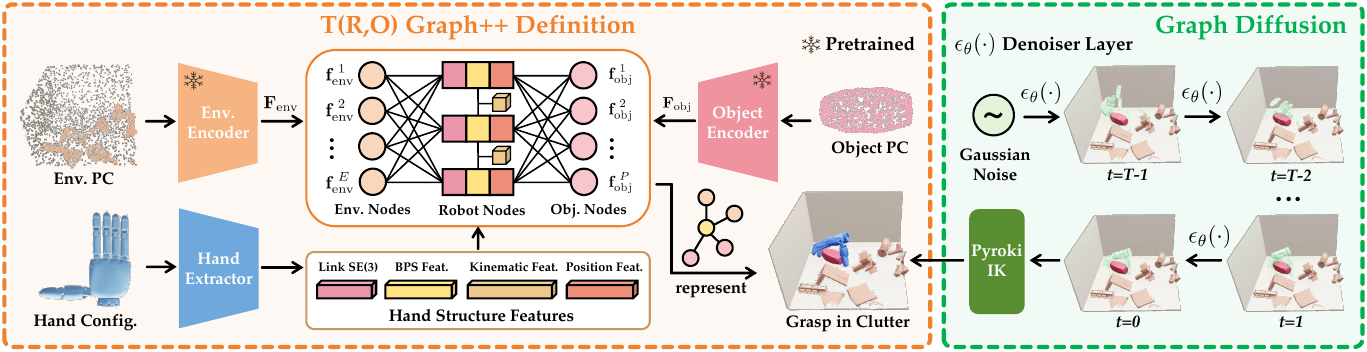}
    \vspace{-15pt}
    \caption{\small \textbf{Overview of $\mathcal{T(R,O)}$ Grasp++ pipeline.}}
    \label{fig:pipeline}
    \vspace{-12pt}
\end{figure}

Given point-cloud observations of the target object and the surrounding environment, denoted as 
$\mathbf{P}_{\text{obj}}$ and $\mathbf{P}_{\text{env}}$, together with the URDF description of an arbitrary dexterous hand, our goal is to learn a unified model that predicts link-wise $\mathrm{SE}(3)$ transformations representing a valid and collision-free grasp in a cluttered environment. Finally, the predicted link poses are converted into executable joint values by solving a real-time inverse-kinematics retargeting problem using Pyroki~\cite{kim2025pyroki}.

Built on $\mathcal{T}(R,O)$ Grasp~\cite{fei2025t}, we propose a diffusion-based framework for cross-embodiment dexterous grasp synthesis in cluttered environments as shown in Fig.~\ref{fig:pipeline}. Our method consists of two key components: 
(1) a unified $\mathcal{T}(R,O)$ Graph++ representation that models hand-object-environment interactions by encoding the kinematics of arbitrary dexterous hands, target-object geometry, and cluttered-scene context as structured graph elements; and 
(2) a scalable diffusion model that performs grasp synthesis based on $\mathcal{T}(R,O)$ Graph++ representation, enabling efficient training and fast inference across diverse dexterous embodiments. We first introduce the unified graph representation across arbitrary embodiments in Sec.~\ref{sec:4.1}. Then, we present the diffusion framework in Sec.~\ref{sec:4.2}. Finally, we describe the training process and data sampling strategy in Sec.~\ref{sec:4.3}.

\subsection{\texorpdfstring{$\mathcal{T}(R,O)$ Graph++}{T(R,O) Graph++}}
\label{sec:4.1}
\vspace{-5pt}
Learning cross-embodiment dexterous grasping in cluttered environments requires a unified representation that captures interactions among arbitrary dexterous hands, target objects, and surrounding environments. Compared with joint values or object-centric contact maps, interaction-based representations~\cite{wei2024dro,fei2025t} have demonstrated better generalization and efficiency. Therefore, we extend $\mathcal{T}(R,O)$ Graph to $\mathcal{T}(R,O)$ Graph++, which jointly encodes object-environment geometry and hand structure. Importantly, each hand link is encoded with both geometry and kinematics as a unique footprint. This design enables the model to distinguish visually similar links across hands for scalable multi-hand training, as validated by our ablation study in Section~\ref{sec:5.5}.

\subsubsection{Object-Environment Geometry Encoding}
\label{sec:4.1.1}
\vspace{-5pt}
To extract geometry for grasping from point clouds, we pretrain a VQ-VAE~\cite{van2017neural} based on PointNet++~\cite{qi2017pointnet++} separately for the target object and the surrounding environment, denoted as $\phi_{\text{obj}}$ and $\phi_{\text{env}}$. Specifically, we normalize both point clouds with the environment centroid and scale, aligning them in a shared normalized coordinate frame. Finally, each patch center is concatenated with its VQ-VAE encoded feature to form the object and environment geometry features:

\vspace{-13pt}
{
\setlength{\abovedisplayskip}{3pt}
\setlength{\belowdisplayskip}{2pt}
\begin{equation}
\label{eq:vqvae}
\begin{aligned}
\mathbf{F}_{\text{obj}}
&=
\operatorname{Concat}\left(
\mathbf{C}_{\text{obj}},
\phi_{\text{obj}}(\mathbf{P}_{\text{obj}})
\right)
\in \mathbb{R}^{P \times (3+d_{\text{obj}})}, \\
\mathbf{F}_{\text{env}}
&=
\operatorname{Concat}\left(
\mathbf{C}_{\text{env}},
\phi_{\text{env}}(\mathbf{P}_{\text{env}})
\right)
\in \mathbb{R}^{E \times (3+d_{\text{env}})},
\end{aligned}
\end{equation}
}
\vspace{-10pt}

where $\mathbf{C}_{\text{obj}}$ and $\mathbf{C}_{\text{env}}$ denote the normalized patch centers of the object and environment point clouds, respectively, while $P$ and $E$ denote the corresponding numbers of patches.

\subsubsection{Hand Structure Encoding}
\label{sec:4.1.2}
\vspace{-5pt}
For large-scale training on multi-hand grasp datasets, link-wise geometry alone is insufficient, as links from different hands may have similar visual geometry but distinct kinematic roles, causing the model to confuse them and produce collapsed predictions. To provide a unique link-level fingerprint across diverse dexterous embodiments, we encode both link geometry and hand kinematics.

Consider a dexterous hand with $L$ links and $\mathrm{DOF}$ joints. To encode link geometry, we adopt a shared Basis Point Set (BPS)~\cite{prokudin2019efficient}. For each link, we sample points from its mesh in the URDF-defined local frame and compute the corresponding BPS feature 
$\mathbf{F}_{\text{robot}}^{\text{BPS}} \in \mathbb{R}^{L \times d_{\text{BPS}}}$. To further capture the hand kinematic structure, we randomly sample $N$ joint configurations from the feasible joint space and perform forward kinematics to obtain the 6D-pose of every link in the hand base frame, denoted as 
$\mathbf{T}_{\text{robot}}^{\text{base}} \in \mathbb{R}^{N \times L \times 6}$. 
For each of the $\mathrm{DOF}$ joints, we compute the relative transformation between its parent and child links across the sampled configurations. We store these kinematic relations as $\mathbf{T}_{\text{robot}}^{\text{rel}} \in \mathbb{R}^{L \times L \times N \times 6}$, where entries are valid only for joint-connected link pairs and zero-padded otherwise. Finally, we encode $\mathbf{T}_{\text{robot}}^{\text{base}}$ and $\mathbf{T}_{\text{robot}}^{\text{rel}}$ into the link position feature $\mathbf{F}_{\text{robot}}^{\text{pos}}$ and kinematic relation feature $\mathbf{F}_{\text{robot}}^{\text{kine}}$ with position encoder $\phi_{\text{pos}}$ and kinematic encoder $\phi_{\text{kine}}$, respectively:
{
\setlength{\abovedisplayskip}{3pt}
\setlength{\belowdisplayskip}{2pt}
\begin{equation}
\label{eq:kinematic}
\mathbf{F}_{\text{robot}}^{\text{pos}}
=
\phi_{\text{pos}}\left(\mathbf{T}_{\text{robot}}^{\text{base}}\right)
\in \mathbb{R}^{L \times d_{\text{pos}}},
\quad
\mathbf{F}_{\text{robot}}^{\text{kine}}
=
\phi_{\text{kine}}\left(\mathbf{T}_{\text{robot}}^{\text{rel}}\right)
\in \mathbb{R}^{L \times L \times d_{\text{kine}}}.
\end{equation}
}

As a result, $\mathbf{F}_{\text{robot}}^{\text{BPS}}$ captures link geometry for grasping, while $\mathbf{F}_{\text{robot}}^{\text{pos}}$ and $\mathbf{F}_{\text{robot}}^{\text{kine}}$ provide kinematic structure to the downstream network. Together, these features form a unique link-level fingerprint, preventing collapsed predictions caused by confusing links with similar visual geometry.

\subsubsection{Graph Construction}
\label{sec:4.1.3}
\vspace{-5pt}
Given object-environment geometry and encoded hand structure, we construct $\mathcal{T}(R,O)$ Graph++ representation based on $\mathrm{SE}(3)$ spatial transformations to model interactions among dexterous embodiments, target objects and surrounding environments. For clarity, we denote the link-wise 6D poses in the world frame as 
$\mathbf{T}_{\text{robot}}^{\text{world}} \in \mathbb{R}^{L \times 6}$. 
For object and environment patches, we define their poses as
$\mathbf{T}_{\text{obj}}^{\text{world}} = [\mathbf{C}_{\text{obj}}, \mathbf{0}] \in \mathbb{R}^{P \times 6}$ and
$\mathbf{T}_{\text{env}}^{\text{world}} = [\mathbf{C}_{\text{env}}, \mathbf{0}] \in \mathbb{R}^{E \times 6}$,
where the rotational components are set to zero since point-cloud patches do not carry canonical orientations.

\textbf{Nodes.} We define object, environment, and robot nodes in the graph. Object and environment nodes are constructed from their encoded geometry features:
$\mathbf{N}_{\text{obj}}=\mathbf{F}_{\text{obj}}$ and 
$\mathbf{N}_{\text{env}}=\mathbf{F}_{\text{env}}$.
Each robot node represents one hand link and is formed by concatenating its 6D-pose in world frame, BPS geometry feature, and kinematic-position feature:
\vspace{-2pt}
{
\setlength{\abovedisplayskip}{3pt}
\setlength{\belowdisplayskip}{-2pt}
\begin{equation}
\label{eq:robot node}
\mathbf{N}_{\text{robot}}
=
\operatorname{Concat}
\left(
\mathbf{T}_{\text{robot}}^{\text{world}},
\mathbf{F}_{\text{robot}}^{\text{BPS}},
\mathbf{F}_{\text{robot}}^{\text{pos}}
\right).
\end{equation}
}

\textbf{Edges.} Since our model predicts link-wise poses, we define object-robot and environment-robot edges to pass scene information to robot-link nodes. Object-robot edges guide each robot link toward grasping the target object, while environment-robot edges provide surrounding clutter information to keep the predicted link poses collision-free with the environment. To encode these spatial relations, we define each edge as the relative 6D pose between its connected nodes:
{
\setlength{\abovedisplayskip}{3pt}
\setlength{\belowdisplayskip}{2pt}
\begin{equation}
\label{eq:obj-robot and env-robot edges}
\mathbf{E}_{\text{or}}
= \operatorname{Rel}
\left(
\mathbf{T}_{\text{obj}}^{\text{world}},
\mathbf{T}_{\text{robot}}^{\text{world}}
\right), \quad
\mathbf{E}_{\text{er}}
= \operatorname{Rel}
\left(
\mathbf{T}_{\text{env}}^{\text{world}},
\mathbf{T}_{\text{robot}}^{\text{world}}
\right),
\end{equation}
}
\vspace{-1pt}
where $\operatorname{Rel}(\cdot)$ denotes the relative 6D-pose operator. We further define robot-robot edges to model inter-link relations and hand kinematics, enabling coordinated pose prediction across hand links. Specifically, each robot-robot edge first encodes the relative 6D pose between a pair of robot links. Then, we augment robot-robot edges with the kinematic relation feature $\mathbf{F}_{\text{robot}}^{\text{kine}}$ to better capture kinematic consistency across hand links:
\vspace{-1pt}
{
\setlength{\abovedisplayskip}{3pt}
\setlength{\belowdisplayskip}{2pt}
\begin{equation}
\label{eq:robot-robot edges}
\mathbf{E}_{\text{rr}}
= \operatorname{Concat}
\left(
\operatorname{Rel}
\left(
\mathbf{T}_{\text{robot}}^{\text{world}},
\mathbf{T}_{\text{robot}}^{\text{world}}
\right),
\mathbf{F}_{\text{robot}}^{\text{kine}}
\right).
\end{equation}
}
\vspace{-1pt}
Under this definition, we encode the object-environment geometry and hand structure, together with their spatial relationships under arbitrary hand embodiments and object-scene configurations, into a unified $\mathcal{T}(R,O)$ Graph++ representation: $\mathcal{G}=(\mathbf{N}_{\text{obj}}, \mathbf{N}_{\text{env}}, \mathbf{N}_{\text{robot}}, \mathbf{E}_{\text{or}}, \mathbf{E}_{\text{er}}, \mathbf{E}_{\text{rr}})$.

\vspace{-2pt}
\subsection{\texorpdfstring{Diffusion Model on $\mathcal{T}(R,O)$ Graph++}{Diffusion Model on T(R,O) Graph++}}
\label{sec:4.2}
\vspace{-5pt}
With the above $\mathcal{T}(R,O)$ Graph++ representation, we reformulate dexterous grasp synthesis in cluttered environments as generating a valid graph that represents a feasible and collision-free grasp. To improve inference efficiency, we adopt DDIM sampling strategy~\cite{song2020denoising}.

\textbf{Forward Process.} Given a graph $\mathcal{G}_0$ representing a valid grasp, we gradually perturb the link-wise 6D poses $\mathbf{T}_{\text{robot}}^{\text{world},0}$ by adding Gaussian noise with variance schedule $\beta_1 < \cdots < \beta_T$. The noisy link poses at diffusion step $t$ are then formulated as:
{
\setlength{\abovedisplayskip}{3pt}
\setlength{\belowdisplayskip}{2pt}
\begin{equation}
\label{eq:forward}
p \left(
\mathbf{T}_{\text{robot}}^{\text{world},t}
\mid
\mathbf{T}_{\text{robot}}^{\text{world},0}
\right)
=
\mathcal{N}
\left(
\sqrt{\bar{\alpha}_t}
\mathbf{T}_{\text{robot}}^{\text{world},0},
(1-\bar{\alpha}_t)\mathbf{I}
\right),
\end{equation}
}
where $\bar{\alpha}_t=\prod_{k=1}^{t}(1-\beta_k)$. Then, we update the corresponding robot nodes and edges as defined in Equations~\ref{eq:robot node}, \ref{eq:obj-robot and env-robot edges}, and \ref{eq:robot-robot edges} to obtain a sequence of noisy graph $\{\mathcal{G}_1, ..., \mathcal{G}_T\}$.

\textbf{Denoiser.} We implement a scalable transformer-based denoiser, denoted as $\boldsymbol{\epsilon}_{\theta}(\mathcal{G}_t,t)$, to predict link-wise pose noise from the noisy $\mathcal{T}(R,O)$ Graph++ representation. At each denoising step, a sinusoidal timestep embedding is injected into all node and edge features. The denoiser then applies $K$ stacked graph-attention layers, where each robot-link node sequentially attends to object nodes, environment nodes, and other robot-link nodes through edge-conditioned attention:
{
\setlength{\abovedisplayskip}{3pt}
\setlength{\belowdisplayskip}{2pt}
\begin{equation}
\label{eq:denoiser}
\mathbf{N}_{\text{robot}}^{\ell+1}
=
\mathcal{A}_{RR}
\left(
\mathcal{A}_{ER}
\left(
\mathcal{A}_{OR}
\left(
\mathbf{N}_{\text{robot}}^{\ell}, \mathbf{N}_{\text{obj}}, \mathbf{E}_{\text{or}}^{\ell}
\right),
\mathbf{N}_{\text{env}}, \mathbf{E}_{er}^{\ell}
\right),
\mathbf{N}_{\text{robot}}^{\ell}, \mathbf{E}_{rr}^{\ell}
\right),
\end{equation}
}
where $\ell=0,\ldots,K-1$ indexes the graph-attention layers, $\mathcal{A}_{OR}$, $\mathcal{A}_{ER}$, and $\mathcal{A}_{RR}$ denote multi-head edge-conditioned attention~\cite{attention} over object-robot, environment-robot, and robot-robot interactions, respectively. Finally, robot-link features from all layers are aggregated and decoded to predict the translational and rotational noise of each link.

\textbf{Reverse Process.} During inference, we initialize link poses from Gaussian noise and use DDIM sampling to progressively denoise the graph. At each reverse step, the denoiser predicts link-wise pose noise $\boldsymbol{\epsilon}_{\theta}(\mathcal{G}_t,t)$, which is used to update link poses to the next DDIM timestep. The corresponding robot nodes and interaction edges are then recomputed according to the updated poses. After the final denoising step, the predicted denoised graph $\hat{\mathcal{G}}_0$ yields the link-wise grasp poses.

\subsection{Training and Data Sampling}
\label{sec:4.3}
\vspace{-5pt}
The denoiser is trained with standard noise-prediction objective in diffusion models. Given a noisy graph $\mathcal{G}_t$ at timestep $t$, the model predicts the injected positional noise $\boldsymbol{\epsilon}^{\text{pos}}$ and rotational noise $\boldsymbol{\epsilon}^{\text{rot}}$. The denoising loss is defined as:
{
\setlength{\abovedisplayskip}{3pt}
\setlength{\belowdisplayskip}{2pt}
\begin{equation}
\label{eq:loss}
\mathcal{L}_{\text{denoise}} =
\gamma_{\text{pos}}
\left\|
\boldsymbol{\epsilon}_{\theta}^{\text{pos}}(\mathcal{G}_t,t)
-
\boldsymbol{\epsilon}^{\text{pos}}
\right\|_2
+
\gamma_{\text{rot}}
\left\|
\boldsymbol{\epsilon}_{\theta}^{\text{rot}}(\mathcal{G}_t,t)
-
\boldsymbol{\epsilon}^{\text{rot}}
\right\|_2 ,
\end{equation}
}
where $\gamma_{\text{pos}}$ and $\gamma_{\text{rot}}$ balance the positional and rotational denoising terms, respectively. During training, we employ a data sampler to maximize the diversity of dexterous embodiments and objects within each batch, reducing hand- and object-specific batch bias and encouraging the model to learn grasping patterns that generalize across different kinematics, objects and clutter configurations.

\section{Experiments}
\label{sec:exp}
\vspace{-5pt}
In this section, we conduct a series of experiments to answer the following questions (Q1-Q6):

\vspace{-5pt}
\begin{itemize}[leftmargin=5pt, itemsep=0pt, topsep=1pt, parsep=0pt, partopsep=0pt]
    \item[] Q1: Can our model perform successful, diverse, and collision-free dexterous grasp synthesis in cluttered environments across different embodiments? (Sec.~\ref{sec:5.2})
    \item[] Q2: How does our model perform across multi-level cluttered environments, and where is its performance boundary under the difficulty metric defined by our benchmark? (Sec.~\ref{sec:5.2})
    \item[] Q3: How efficient is our proposed model in inference and sampling compared with single-object frameworks and scene-level proposal-and-selection methods? (Sec.~\ref{sec:5.3})
    \item[] Q4: What is the scaling law for dexterous grasping synthesis? (Sec.~\ref{sec:5.4})
    \item[] Q5: How important is the link footprint for scalable cross-embodiment learning? (Sec.~\ref{sec:5.5})
    \item[] Q6: How does our model perform in real-world settings? (Sec.~\ref{sec:5.6})
\end{itemize}

\subsection{Experimental Setting}
\label{sec:5.1}
\vspace{-5pt}
We compare our method with several baselines on the test scenes described in Sec.~\ref{sec:3.1}. Following the evaluation protocol in Sec.~\ref{sec:3.4}, we evaluate all generated grasps in Isaac Gym simulation and report both success rate and grasp diversity across multiple dexterous hands. Due to limited available data and hand-specific hard-coded patterns in existing baselines, we conduct comparisons only on embodiments shared by all methods. Since AnyDexGrasp~\cite{fang2025anydexgrasp} produces grasp proposals for all objects in a scene, we first filter the proposals associated with the target object and then select the highest-scoring one as its prediction. For DexGraspNet2~\cite{zhang2024dexgraspnet}, we similarly select the highest-scoring grasp from all generated candidates. Detailed implementation details are presented in Appendix.

\subsection{Cross-Embodiment Grasp Synthesis in Cluttered Environment}
\label{sec:5.2}
\vspace{-5pt}
\begin{table*}[t]
    \centering
    \scriptsize
    \renewcommand\arraystretch{1.2}
    \setlength{\tabcolsep}{4pt}
    \resizebox{\textwidth}{!}{
        \begin{tabular}{c|cccccccc|c}
            \toprule
            \multirow{2}{*}[-0.2ex]{\textbf{Method}}
            & \multicolumn{9}{c}{\textbf{Success Rate (\%) $\uparrow$ / Diversity (rad.) $\uparrow$}}
            \\
            \cline{2-10}
            & Allegro
            & Ability
            & Barrett
            & DClaw
            & LEAP
            & Schunk
            & Shadow
            & XHand
            & Avg.
            \\
            \midrule
            AnyDexGrasp~\cite{fang2025anydexgrasp}
                & 54.44/\textbf{0.387} & -- & -- & -- & -- & -- & -- & -- & 54.44/0.387
            \\
            DexGraspNet2~\cite{zhang2024dexgraspnet}
                & -- & -- & -- & -- & 52.09/0.303 & -- & -- & -- & 52.09/0.303
            \\
            \textbf{Ours (Single-trial)}
                & \textbf{88.79}/0.369
                & \textbf{88.28}/\textbf{0.509} 
                & \textbf{87.24}/\textbf{0.470} 
                & \textbf{89.11}/\textbf{0.393} 
                & \textbf{85.93}/\textbf{0.404}
                & \textbf{90.39}/\textbf{0.389} 
                & \textbf{87.87}/\textbf{0.354} 
                & \textbf{86.03}/\textbf{0.418} 
                & \textbf{87.96}/\textbf{0.413}
            \\
            \bottomrule
        \end{tabular}
    }
    \vspace{-5pt}
    \caption{\small \textbf{Overall success rate and diversity comparison across different dexterous hands.}}
    \label{tab:main_result}
    \vspace{-8pt}
\end{table*}

Based on the 5,938 unseen objects as described in Sec.~\ref{sec:3.1}, we generate 20,000 test scenes for each dexterous hand for evaluation. As shown in Tab.~\ref{tab:main_result}, our method significantly outperforms both baselines, achieving an average success rate of 88.0\% across multiple dexterous embodiments. Fig.~\ref{fig:vis_with_baseline} illustrates that AnyDexGrasp often over-emphasizes fitting grasp templates to the target object, while neglecting environmental collision constraints. DexGraspNet2 leverages local geometric features for cluttered-scene grasp generation, but its selected candidates still fail to achieve accurate target-object contacts while avoiding scene collisions. In addition, AnyDexGrasp relies on hard-coded, hand-specific grasp templates, while DexGraspNet2 directly predicts hand joint values. As a result, neither method supports cross-embodiment dexterous grasp synthesis in cluttered environments.

To further explore the performance boundary in highly cluttered environments, we generate 20,000 challenging test scenes on unseen objects with denser clutter and a larger number of obstacle objects. Combining them with the original test set, we obtain 40,000 test scenes for each hand and split them into four difficulty levels according to $D_{\text{env}}$. The score range of each level is shown in Tab.~\ref{tab:hierarchical_result}. For comparison, we report the average success rate and diversity across all hands for our method, and over the supported shared hands for each baseline. As shown in Tab.~\ref{tab:hierarchical_result}, our method consistently outperforms both baselines across all difficulty levels. Its performance gradually decreases as $D_{\text{env}}$ increases, but remains above 75\% success rate even in Level-4 clutter, indicating a robust performance boundary in highly constrained scenes.

\vspace{-5pt}

\begin{table*}[t]
    \centering
    \scriptsize
    \renewcommand\arraystretch{1.15}
    \setlength{\tabcolsep}{2.2pt}

    \begin{minipage}[t]{0.49\textwidth}
        \vspace{0pt}
        \centering
        \scriptsize
        \renewcommand\arraystretch{1.15}
        \setlength{\tabcolsep}{3pt}
        \resizebox{\linewidth}{!}{
            \begin{tabular}{c|cccc}
                \toprule
                \multirow{2}{*}[-0.2ex]{\textbf{Method}}
                & \multicolumn{4}{c}{\textbf{Success Rate (\%) $\uparrow$ / Diversity (rad.) $\uparrow$}}
                \\
                \cline{2-5}
                & \makecell{\rule[-1.0ex]{0pt}{3.5ex}Level-1\\$\in$[0, 0.5)}
                & \makecell{\rule[-1.0ex]{0pt}{3.5ex}Level-2\\$\in$[0.5, 0.75)}
                & \makecell{\rule[-1.0ex]{0pt}{3.5ex}Level-3\\$\in$[0.75, 0.9)}
                & \makecell{\rule[-1.0ex]{0pt}{3.5ex}Level-4\\$\in$[0.9, 1.0]}
                \\
                \midrule
                AnyDexGrasp~\cite{fang2025anydexgrasp}
                    & 53.09/0.306
                    & 50.82/0.298
                    & 48.73/0.281
                    & 44.27/0.282
                \\
                DexGraspNet2~\cite{zhang2024dexgraspnet}
                    & 55.28/0.308
                    & 54.76/0.311
                    & 51.83/0.315
                    & 50.56/0.310
                \\
                \textbf{Ours}
                    & \textbf{80.32}/\textbf{0.408}
                    & \textbf{77.72}/\textbf{0.418}
                    & \textbf{77.81}/\textbf{0.423}
                    & \textbf{75.37}/\textbf{0.412}
                \\
                \bottomrule
            \end{tabular}
        }
        \vspace{-3pt}
        \caption{\small \textbf{Hierarchical comparisons.}}
        \label{tab:hierarchical_result}
    \end{minipage}%
    \hfill%
    \begin{minipage}[t]{0.49\textwidth}
        \vspace{0pt}
        \centering
        \scriptsize
        \renewcommand\arraystretch{1.15}
        \setlength{\tabcolsep}{3pt}
        \resizebox{\linewidth}{!}{
            \begin{tabular}{c|cccc}
                \toprule
                \multirow{2}{*}[-0.2ex]{\textbf{Method}}
                & \multicolumn{4}{c}{\textbf{Efficiency Metrics}}
                \\
                \cline{2-5}
                & Params (M)
                & Latency (s) $\downarrow$
                & Throughput $\uparrow$
                & PPS $\downarrow$
                \\
                \midrule
                $\mathcal{T(R,O)}$ Grasp~\cite{fei2025t}
                    & 28.06 & \textbf{0.210} & 16.97 & $>$100
                \\
                AnyDexGrasp~\cite{fang2025anydexgrasp}
                    & 49.93 & 0.835 & 1.28 & 26.39
                \\
                DexGraspNet2~\cite{zhang2024dexgraspnet}
                    & 15.10 & 1.712 & 4.96 & 13.34
                \\
                \textbf{Ours}
                    & 176.53 & 0.336 & \textbf{65.31} & \textbf{1.15}
                \\
                \bottomrule
            \end{tabular}
        }
        \vspace{-3pt}
        \caption{\small\textbf{Efficiency analysis.}}
        \label{tab:efficiency}
    \end{minipage}
    \vspace{-15pt}
\end{table*}

\subsection{Efficiency Analysis}
\label{sec:5.3}
\vspace{-5pt}
In this section, we evaluate both the inference efficiency and sample efficiency of our proposed framework. For inference efficiency, we compare the inference latency and throughput of all methods on an NVIDIA 4090 GPU. To measure sample efficiency during inference, we define \textbf{Proposals Per Success (PPS)}, which measures the number of grasp proposals a model needs to generate to obtain one successful grasp. For AnyDexGrasp and DexGraspNet2, we first extract target-object grasp proposals from all scene-level proposals, and then evaluate them to compute PPS. As shown in Tab.~\ref{tab:efficiency}, our model achieves an inference speed of around 3 FPS with 0.2B model parameters. In addition, compared with proposal-and-selection and single-object grasp synthesis methods, including AnyDexGrasp, DexGraspNet2 and $\mathcal{T(R,O)}$ Grasp, our method achieves substantially better sample efficiency during inference. These results show that our framework enables efficient grasp synthesis and can be practically deployed on a single consumer-grade GPU.

\subsection{Scaling Law and Multi-Trial Analysis}
\label{sec:5.4}
\vspace{-5pt}

\begin{figure*}[!t]
    \centering

    \begin{minipage}[t]{\columnwidth}
        \vspace{0pt}
        \centering
        \includegraphics[width=\linewidth]{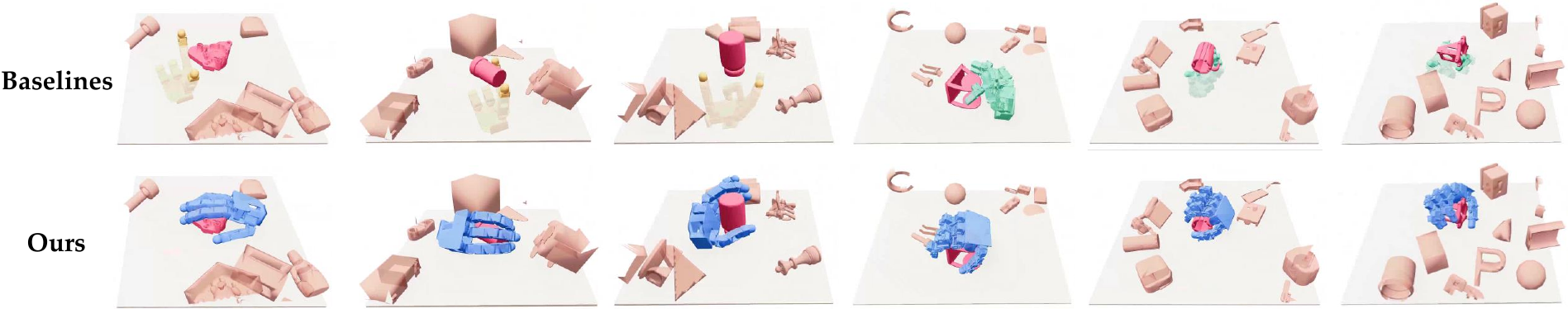}
        \vspace{-15pt}
        \captionof{figure}{\small \textbf{Visualization comparisons with baselines.} Compared with baselines, $\mathcal{T}(R,O)$ Grasp++ performs accurate and collision-free grasp synthesis. Red/brown denote target/obstacle objects.}
        \label{fig:vis_with_baseline}
    \end{minipage}%
    \vspace{8pt}
    \begin{minipage}[t]{\columnwidth}
        \vspace{0pt}
        \centering
        \includegraphics[width=\linewidth]{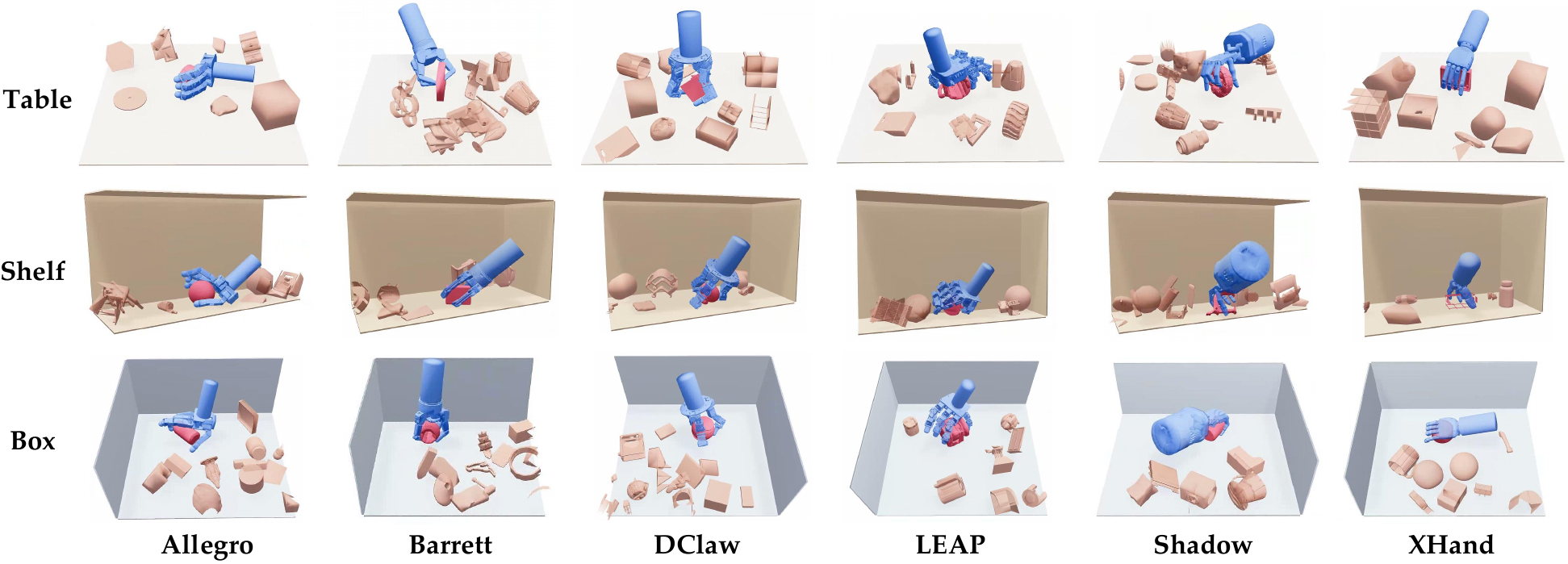}
        \vspace{-15pt}
        \captionof{figure}{\small \textbf{Cross-embodiment grasp synthesis in clutter.} $\mathcal{T}(R,O)$ Grasp++ achieves cross-embodiment grasp synthesis in common manipulation environments, including tabletop, shelf, and box.}
        \label{fig:multi_hand_res}
    \end{minipage}

    \vspace{10pt}
    \begin{minipage}[t]{0.53\textwidth}
        \vspace{0pt}
        \centering
        \includegraphics[width=\linewidth]{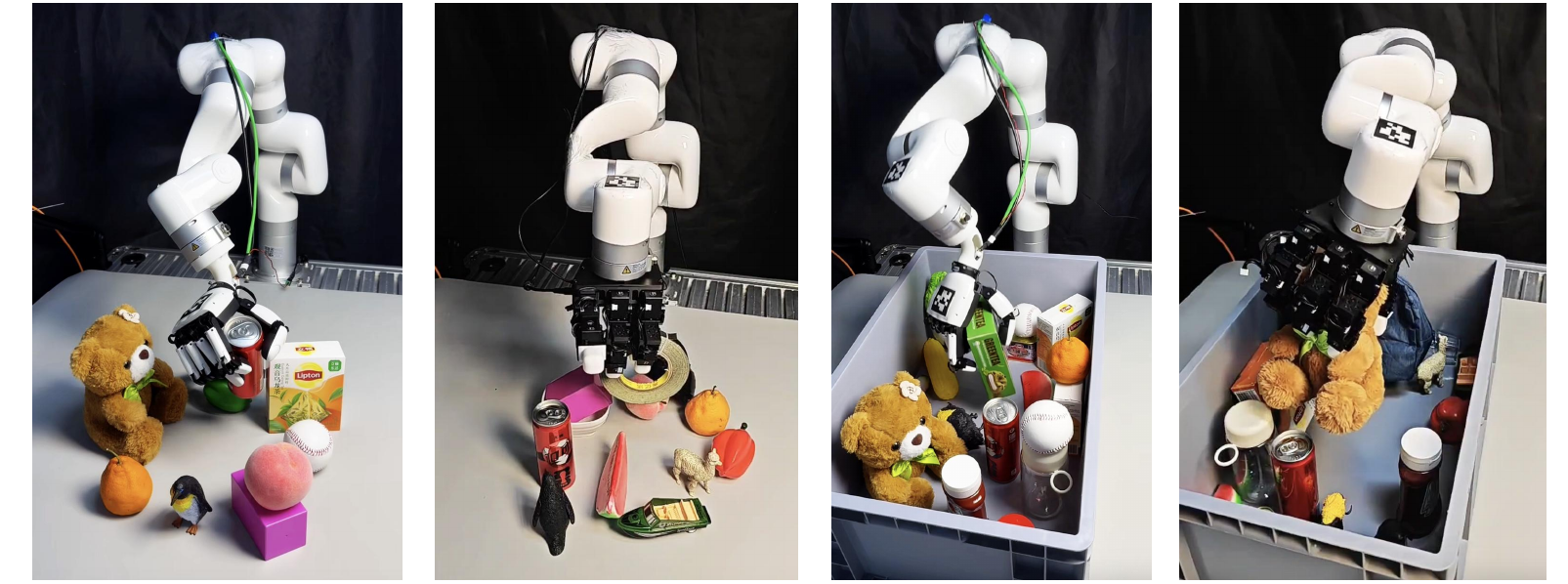}
        \vspace{-15pt}
        \captionof{figure}{\small \textbf{Real-world settings.}}
        \label{fig:real_world}
    \end{minipage}%
    \begin{minipage}[t]{0.47\textwidth}
        \vspace{0pt}
        \centering
        \scriptsize
        \renewcommand\arraystretch{1.15}
        \setlength{\tabcolsep}{3.4pt}

        \resizebox{\linewidth}{!}{
        \begin{tabular}{@{}cc|cc|cc@{}}
            \toprule
            \multicolumn{2}{c|}{\textbf{Model Size}} &
            \multicolumn{2}{c|}{\textbf{Object Scale}} &
            \multicolumn{2}{c}{\textbf{Multi-Trial}} \\
            \midrule
            Params & SR (\%) $\uparrow$ &
            Objects & SR (\%) $\uparrow$ &
            Trials & SR (\%) $\uparrow$ \\
            \midrule
            10M  & 79.63 &
            12K  & 83.65 &
            1    & 87.96 \\
            50M  & 86.77 &
            24K  & 84.15 &
            2    & 95.58 \\
            200M & 87.96 &
            36K  & 85.29 &
            3    & 97.61 \\
            1B   & \textbf{88.07} &
            48K  & \textbf{86.24} &
            4    & \textbf{98.46} \\
            \bottomrule
        \end{tabular}
        }

        \vspace{-2pt}
        \captionof{table}{\small \textbf{Scaling and multi-trial analysis.}}
        \label{tab:scaling_multi_trial}
    \end{minipage}

    \vspace{-10pt}
\end{figure*}

In this section, we first investigate the scaling laws of our method along model size and training object scale. As shown in Tab.~\ref{tab:scaling_multi_trial}, scaling the model up to 200M parameters consistently improves performance, highlighting the benefit of increased capacity. Further scaling to 1B parameters brings marginal gains, suggesting performance saturation beyond 200M under the current training setting and data scale. Along another axis, increasing the number of training objects consistently improves the success rate. This suggests that training on more objects exposes the model to a wider range of successful grasp patterns, improving its generalization to unseen object geometries.

We further evaluate our method under a multi-trial evaluation setting, where each scene is tested with multiple independent trials and considered successful if any trial successfully lifts up the target object. As shown in Tab.~\ref{tab:scaling_multi_trial}, the success rate improves from 87.96\% with one trial to 98.46\% with four trials, suggesting that multiple samples can recover feasible grasps missed by single-trial sampling. Such failures may be further reduced with execution feedback and closed-loop strategies.

\subsection{Ablation}
\label{sec:5.5}

As discussed in Sec.~\ref{sec:4.1}, $\mathcal{T}(R,O)$ Graph++ encodes each robot link with a unique footprint that combines local link geometry and kinematic structure. This footprint allows the model to distinguish visually similar links with different kinematic roles across dexterous hands. To validate this design, we conduct an ablation study by removing the link-footprint encoding, including positional feature $\mathbf{F}_{\text{robot}}^{\text{pos}}$ and kinematic feature $\mathbf{F}_{\text{robot}}^{\text{kine}}$, while retaining only geometry feature $\mathbf{F}_{\text{robot}}^{\text{BPS}}$. We then evaluate how this geometry-only representation affects cross-embodiment grasp synthesis.

\begin{figure}[t]
    \centering

    \begin{minipage}[t]{0.60\linewidth}
        \vspace{0pt}
        \centering
        \scriptsize
        \renewcommand{\arraystretch}{1.05}
        \setlength{\tabcolsep}{3pt}

        \newcommand{\tabrow}{\rule{0pt}{2.4ex}}

        \resizebox{\linewidth}{!}{
            \begin{tabular}{l|cccc|c}
                \toprule
                \tabrow
                \textbf{Setting}
                & \textbf{Allegro}
                & \textbf{LEAP}
                & \textbf{Shadow}
                & \textbf{XHand}
                & \textbf{Avg.}
                \\
                \midrule
                \tabrow
                Model w/o Link-Footprint
                & 23.38
                & 4.31
                & 4.84
                & 5.91
                & 9.61
                \\
                \tabrow
                Full Model
                & \textbf{88.79}
                & \textbf{85.93}
                & \textbf{87.87}
                & \textbf{86.03}
                & \textbf{87.16}
                \\
                \bottomrule
            \end{tabular}
        }

        \vspace{-3pt}
        \captionof{table}{\small \textbf{Ablation of link-footprint encoding.}}
        \label{tab:link_footprint_ablation}
    \end{minipage}
    \hfill
    \begin{minipage}[t]{0.38\linewidth}
        \vspace{0pt}
        \centering
        \includegraphics[width=\linewidth]{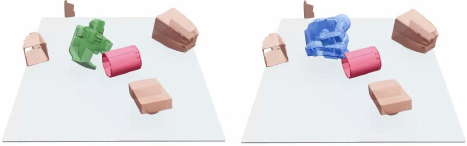}

        \vspace{-3pt}
        \captionof{figure}{\small \textbf{Mode collapse.}}
        \label{fig:link_footprint_ablation}
    \end{minipage}

\end{figure}

As shown in Tab.~\ref{tab:link_footprint_ablation}, removing the positional and kinematic encodings causes severe performance degradation and model collapse. On Allegro Hand, multiple finger links share highly similar geometries, making them difficult to distinguish using only the geometry feature $\mathbf{F}_{\text{robot}}^{\text{BPS}}$. As illustrated in Fig.~\ref{fig:link_footprint_ablation}, this ambiguity causes the predicted link $\text{SE(3)}$ poses to collapse despite their different kinematic positions and functional roles. These results demonstrate that link-footprint encoding is crucial for preserving distinctive link identities and supporting scalable multi-hand learning.

\subsection{Real-world Experiments}
\label{sec:5.6}
\vspace{-5pt}
We conduct real-world experiments using a uFactory xArm7 robot equipped with an XHand and a LEAP Hand~\cite{shaw2023leap}, as illustrated in Fig.~\ref{fig:real_world}. Object observations are captured by three Intel RealSense depth cameras covering the manipulation workspace. We evaluate $\mathcal{T}(R,O)$ Grasp++ on 10 novel daily objects in cluttered tabletop and in-box scenarios, where the target object is surrounded by obstacle objects to introduce occlusion and physical interference. Our method achieves success rates of \textbf{85\%} and \textbf{82\%} on \textbf{XHand} and \textbf{LEAP Hand}, respectively, demonstrating its effectiveness in real-world cluttered environments.

We further extend our framework to real-world grasping from single-view observations. Single-view training data are synthesized by projecting full point clouds from randomized camera viewpoints and removing occluded points. For each embodiment used in single-view evaluation, we generate approximately 6 million grasp annotations and fine-tune the pretrained model to convergence. Across more than 40 real-world objects with approximately 10 trials per object, our method achieves success rates of \textbf{82.1\%}, \textbf{90.6\%}, and \textbf{98.5\%} on \textbf{Wuji}, \textbf{XHandPro}, and the \textbf{xArm Gripper}, respectively. Further details on the real-world evaluation and qualitative results are provided in Sec.~\ref{sec:7.1}, with videos available on our project webpage.
\section{Limitations and Conclusions}
\label{sec:conclusion}
\vspace{-5pt}
In this paper, we introduce $\mathcal{T}(R,O)$ Grasp++, a scalable framework for cross-embodiment grasping synthesis in clutter, supported by a large-scale benchmark containing around 0.5B grasp annotations. Extensive experiments demonstrate that $\mathcal{T}(R,O)$ Grasp++ outperforms existing baselines in simulation and remains robust in real-world cluttered scenes. We also validate its inference and sample efficiency, and analyze scaling trends for future research directions.

However, our work has several limitations. First, $\mathcal{T}(R,O)$ Grasp++ does not support multi-modal inputs, such as RGB channels, force feedback, or tactile signals. As a result, our model cannot leverage appearance cues or contact feedback to refine grasp synthesis. Second, our data generation pipeline has limited capability for functional grasp annotation. Therefore, $\mathcal{T}(R,O)$ Grasp++ does not yet support functional grasp synthesis. We hope future work can address these limitations.







\clearpage

\section*{Acknowledgement}

This research is supported by the Ministry of Education, Singapore, under the Academic Research Fund Tier 1 (FY2024), and by the National Research Foundation, Prime Minister’s Office, Singapore under its Campus for Research Excellence and Technological Enterprise (CREATE) programme.

\bibliography{ref}  

@inproceedings{xu2023unidexgrasp,
  title={Unidexgrasp: Universal robotic dexterous grasping via learning diverse proposal generation and goal-conditioned policy},
  author={Xu, Yinzhen and Wan, Weikang and Zhang, Jialiang and Liu, Haoran and Shan, Zikang and Shen, Hao and Wang, Ruicheng and Geng, Haoran and Weng, Yijia and Chen, Jiayi and others},
  booktitle={CVPR},
  pages={4737--4746},
  year={2023}
}

@article{deitke2023objaverse,
  title={Objaverse-xl: A universe of 10m+ 3d objects},
  author={Deitke, Matt and Liu, Ruoshi and Wallingford, Matthew and Ngo, Huong and Michel, Oscar and Kusupati, Aditya and Fan, Alan and Laforte, Christian and Voleti, Vikram and Gadre, Samir Yitzhak and others},
  journal={NeurIPS},
  volume={36},
  pages={35799--35813},
  year={2023}
}

@article{fei2025t,
  title={T (R, O) Grasp: Efficient Graph Diffusion of Robot-Object Spatial Transformation for Cross-Embodiment Dexterous Grasping},
  author={Fei, Xin and Xu, Zhixuan and Fang, Huaicong and Zhang, Tianrui and Shao, Lin},
  journal={ICRA},
  year={2026}
}

@article{wei2024dro,
    title={D(R,O) Grasp: A Unified Representation of Robot and Object Interaction for Cross-Embodiment Dexterous Grasping},
    author={Wei, Zhenyu and Xu, Zhixuan and Guo, Jingxiang and Hou, Yiwen and Gao, Chongkai and Cai, Zhehao and Luo, Jiayu and Shao, Lin},
    journal={ICRA},
    year={2025}
}

@inproceedings{kim2025pyroki,
  title={Pyroki: A modular toolkit for robot kinematic optimization},
  author={Kim, Chung Min and Yi, Brent and Choi, Hongsuk and Ma, Yi and Goldberg, Ken and Kanazawa, Angjoo},
  booktitle={IROS},
  pages={1312--1319},
  year={2025},
  organization={IEEE}
}

@article{qi2017pointnet++,
  title={Pointnet++: Deep hierarchical feature learning on point sets in a metric space},
  author={Qi, Charles Ruizhongtai and Yi, Li and Su, Hao and Guibas, Leonidas J},
  journal={NeurIPS},
  volume={30},
  year={2017}
}

@article{van2017neural,
  title={Neural discrete representation learning},
  author={Van Den Oord, Aaron and Vinyals, Oriol and others},
  journal={NeurIPS},
  volume={30},
  year={2017}
}

@inproceedings{prokudin2019efficient,
  title={Efficient learning on point clouds with basis point sets},
  author={Prokudin, Sergey and Lassner, Christoph and Romero, Javier},
  booktitle={ICCV},
  pages={4332--4341},
  year={2019}
}

@article{song2020denoising,
  title={Denoising diffusion implicit models},
  author={Song, Jiaming and Meng, Chenlin and Ermon, Stefano},
  journal={arXiv preprint arXiv:2010.02502},
  year={2020}
}

@inproceedings{zhang2024dexgraspnet,
  title={DexGraspNet 2.0: Learning Generative Dexterous Grasping in Large-scale Synthetic Cluttered Scenes},
  author={Zhang, Jialiang and Liu, Haoran and Li, Danshi and Yu, XinQiang and Geng, Haoran and Ding, Yufei and Chen, Jiayi and Wang, He},
  booktitle={CoRL},
  year={2024}
}

@article{fang2025anydexgrasp,
  title={AnyDexGrasp: General Dexterous Grasping for Different Hands with Human-level Learning Efficiency},
  author={Fang, Hao-Shu and Yan, Hengxu and Tang, Zhenyu and Fang, Hongjie and Wang, Chenxi and Lu, Cewu},
  journal={arXiv preprint arXiv:2502.16420},
  year={2025}
}

@inproceedings{jiang2021graspTTA,
  title={Hand-Object Contact Consistency Reasoning for Human Grasps Generation},
  author={Jiang, Hanwen and Liu, Shaowei and Wang, Jiashun and Wang, Xiaolong},
  booktitle={CVPR},
  year={2021}
}

@inproceedings{casas2024multigrippergrasp,
  title={Multigrippergrasp: A dataset for robotic grasping from parallel jaw grippers to dexterous hands},
  author={Casas, Luis Felipe and Khargonkar, Ninad and Prabhakaran, Balakrishnan and Xiang, Yu},
  booktitle={IROS},
  pages={2978--2984},
  year={2024},
  organization={IEEE}
}

@inProceedings{zhang2024graspxl,
  title={{GraspXL}: Generating Grasping Motions for Diverse Objects at Scale},
  author={Zhang, Hui and Christen, Sammy and Fan, Zicong and Hilliges, Otmar and Song, Jie},
  booktitle={ECCV},
  year={2024}
}

@inproceedings{zhang2025contactdexnet,
  title={Contactdexnet: Multi-fingered robotic hand grasping in cluttered environments through hand-object contact semantic mapping},
  author={Zhang, Lei and Bai, Kaixin and Huang, Guowen and Bing, Zhenshan and Chen, Zhaopeng and Knoll, Alois and Zhang, Jianwei},
  booktitle={IROS},
  pages={8495--8502},
  year={2025},
  organization={IEEE}
}

@inproceedings{wan2023unidexgrasp++,
  title={Unidexgrasp++: Improving dexterous grasping policy learning via geometry-aware curriculum and iterative generalist-specialist learning},
  author={Wan, Weikang and Geng, Haoran and Liu, Yun and Shan, Zikang and Yang, Yaodong and Yi, Li and Wang, He},
  booktitle={CVPR},
  pages={3891--3902},
  year={2023}
}

@article{shao2020unigrasp,
  title={Unigrasp: Learning a unified model to grasp with multifingered robotic hands},
  author={Shao, Lin and Ferreira, Fabio and Jorda, Mikael and Nambiar, Varun and Luo, Jianlan and Solowjow, Eugen and Ojea, Juan Aparicio and Khatib, Oussama and Bohg, Jeannette},
  journal={IEEE Robotics and Automation Letters},
  volume={5},
  number={2},
  pages={2286--2293},
  year={2020},
  publisher={IEEE}
}

@misc{xu2024manifoundation,
      title={ManiFoundation Model for General-Purpose Robotic Manipulation of Contact Synthesis with Arbitrary Objects and Robots}, 
      author={Zhixuan Xu and Chongkai Gao and Zixuan Liu and Gang Yang and Chenrui Tie and Haozhuo Zheng and Haoyu Zhou and Weikun Peng and Debang Wang and Tianyi Chen and Zhouliang Yu and Lin Shao},
      year={2024},
      eprint={2405.06964},
      archivePrefix={arXiv},
      primaryClass={cs.RO}
}

@article{dfc,
  title={Synthesizing diverse and physically stable grasps with arbitrary hand structures using differentiable force closure estimator},
  author={Liu, Tengyu and Liu, Zeyu and Jiao, Ziyuan and Zhu, Yixin and Zhu, Song-Chun},
  journal={IEEE Robotics and Automation Letters},
  volume={7},
  number={1},
  pages={470--477},
  year={2021},
  publisher={IEEE}
}

@inproceedings{li2023gendexgrasp,
  title={Gendexgrasp: Generalizable dexterous grasping},
  author={Li, Puhao and Liu, Tengyu and Li, Yuyang and Geng, Yiran and Zhu, Yixin and Yang, Yaodong and Huang, Siyuan},
  booktitle={ICRA},
  pages={8068--8074},
  year={2023},
  organization={IEEE}
}

@InProceedings{pmlr-v205-chen23b,
  title={Learning Robust Real-World Dexterous Grasping Policies via Implicit Shape Augmentation},
  author={Chen, Qiuyu and Wyk, Karl Van and Chao, Yu-Wei and Yang, Wei and Mousavian, Arsalan and Gupta, Abhishek and Fox, Dieter},
  booktitle={CoRL},
  pages={1222--1232},
  year={2023}
}

@INPROCEEDINGS{hgcnet,
  author={Li, Yiming and Wei, Wei and Li, Daheng and Wang, Peng and Li, Wanyi and Zhong, Jun},
  booktitle={ICRA}, 
  title={HGC-Net: Deep Anthropomorphic Hand Grasping in Clutter}, 
  year={2022},
  pages={714-720}
}

@inproceedings{ye2025dex1b,
    title   = {Dex1B: Learning with 1B Demonstrations for Dexterous Manipulation},
    author  = {Ye, Jianglong and Wang, Keyi and Yuan, Chengjing and Yang, Ruihan and Li, Yiquan and Zhu, Jiyue and Qin, Yuzhe and Zou, Xueyan and Wang, Xiaolong},
    booktitle = {RSS},
    year    = {2025}
}

@ARTICLE{DDGC,
  author={Lundell, Jens and Verdoja, Francesco and Kyrki, Ville},
  journal={IEEE Robotics and Automation Letters}, 
  title={DDGC: Generative Deep Dexterous Grasping in Clutter}, 
  year={2021}
}

@inproceedings{multimodal2021,
    author = {Corsaro, Matt and Tellex, Stefanie and Konidaris, George},
    title = {Learning to Detect Multi-Modal Grasps for Dexterous Grasping in Dense Clutter},
    year = {2021},
    booktitle = {IROS},
}

@inproceedings{zhang2025cadgrasp,
  title={CADGrasp: Learning Contact and Collision Aware General Dexterous Grasping in Cluttered Scenes},
  author={Zhang, Jiyao and Ma, Zhiyuan and Wu, Tianhao and Chen, Zeyuan and Dong, Hao},
  booktitle={NeurIPS},
  year={2025}
}

@article{wang2022dexgraspnet,
  title={Dexgraspnet: A large-scale robotic dexterous grasp dataset for general objects based on simulation},
  author={Wang, Ruicheng and Zhang, Jialiang and Chen, Jiayi and Xu, Yinzhen and Li, Puhao and Liu, Tengyu and Wang, He},
  journal={arXiv preprint arXiv:2210.02697},
  year={2022}
}

@inproceedings{chen2025clutterdexgrasp,
  title={ClutterDexGrasp: A Sim-to-Real System for General Dexterous Grasping in Cluttered Scenes},
  author={Chen, Zeyuan and Yan, Qiyang and Chen, Yuanpei and Wu, Tianhao and Zhang, Jiyao and Ding, Zihan and Li, Jinzhou and Yang, Yaodong and Dong, Hao},
  booktitle={CoRL},
  year={2025}
}

@InProceedings{pmlr-v270-wang25l,
  title={Neural Attention Field: Emerging Point Relevance in 3D Scenes for One-Shot Dexterous Grasping},
  author={Wang, Qianxu and Deng, Congyue and Lum, Tyler Ga Wei and Chen, Yuanpei and Yang, Yaodong and Bohg, Jeannette and Zhu, Yixin and Guibas, Leonidas},
  booktitle={CoRL},
  pages={4495--4508},
  year={2025}
}

@inproceedings{attarian2023geometry,
  title={Geometry Matching for Multi-Embodiment Grasping},
  author={Attarian, Maria and Asif, Muhammad Adil and Liu, Jingzhou and Hari, Ruthrash and Garg, Animesh and Gilitschenski, Igor and Tompson, Jonathan},
  booktitle={CoRL},
  year={2023}
}

@inproceedings{khargonkar2023neuralgrasps,
  title={Neuralgrasps: Learning implicit representations for grasps of multiple robotic hands},
  author={Khargonkar, Ninad and Song, Neil and Xu, Zesheng and Prabhakaran, Balakrishnan and Xiang, Yu},
  booktitle={CoRL},
  pages={516--526},
  year={2023},
  organization={PMLR}
}

@InProceedings{pmlr-v155-breyer21a,
  title={Volumetric Grasping Network: Real-time 6 DOF Grasp Detection in Clutter},
  author={Breyer, Michel and Chung, Jen Jen and Ott, Lionel and Siegwart, Roland and Nieto, Juan},
  booktitle={CoRL},
  pages={1602--1611},
  year={2021}
}

@InProceedings{pmlr-v205-wu23b,
  title={Learning Diverse and Physically Feasible Dexterous Grasps with Generative Model and Bilevel Optimization},
  author={Wu, Albert and Guo, Michelle and Liu, Karen},
  booktitle={CoRL},
  pages={1938--1948},
  year={2023}
}

@article{shaw2023leap,
  title={Leap hand: Low-cost, efficient, and anthropomorphic hand for robot learning},
  author={Shaw, Kenneth and Agarwal, Ananye and Pathak, Deepak},
  journal={arXiv preprint arXiv:2309.06440},
  year={2023}
}

@inproceedings{attention,
    author = {Vaswani, Ashish and Shazeer, Noam and Parmar, Niki and Uszkoreit, Jakob and Jones, Llion and Gomez, Aidan N. and Kaiser, \L{}ukasz and Polosukhin, Illia},
    title = {Attention is all you need},
    year = {2017},
    booktitle = {NeurIPS},
}

@inproceedings{kirillov2023segment,
  title={Segment anything},
  author={Kirillov, Alexander and Mintun, Eric and Ravi, Nikhila and Mao, Hanzi and Rolland, Chloe and Gustafson, Laura and Xiao, Tete and Whitehead, Spencer and Berg, Alexander C and Lo, Wan-Yen and others},
  booktitle={CVPR},
  pages={4015--4026},
  year={2023}
}

@article{schulman2017proximal,
  title={Proximal policy optimization algorithms},
  author={Schulman, John and Wolski, Filip and Dhariwal, Prafulla and Radford, Alec and Klimov, Oleg},
  journal={arXiv preprint arXiv:1707.06347},
  year={2017}
}

@article{loshchilov2017decoupled,
  title={Decoupled weight decay regularization},
  author={Loshchilov, Ilya and Hutter, Frank},
  journal={arXiv preprint arXiv:1711.05101},
  year={2017}
}

@article{chen2024bodex,
  title={BODex: Scalable and Efficient Robotic Dexterous Grasp Synthesis Using Bilevel Optimization},
  author={Chen, Jiayi and others},
  journal={ICRA},
  year={2025}
}
\clearpage
\section*{Appendix}

\section{Real-World Experiment}
\label{sec:7.1}

\paragraph{Real-World Deployment Details.}
We first calibrate the intrinsic and extrinsic parameters of the Intel RealSense depth cameras to transform captured point clouds into a common world coordinate frame. Depending on the evaluation setting, we either fuse observations from multiple viewpoints into a unified scene-level point cloud or directly use a single-view observation. SAM~\cite{kirillov2023segment} is then applied to segment the target object, after which the target-object and surrounding-scene point clouds are extracted and downsampled to match the input format of $\mathcal{T}(R,O)$ Grasp++. We evaluate both XHand and LEAP Hand~\cite{shaw2023leap} in cluttered in-box and tabletop settings, with five objects per setting and randomly placed obstacles around the target. After grasp inference, Pyroki~\cite{kim2025pyroki} is used for motion planning and collision avoidance, enabling the xArm7 robot to reach the predicted grasp pose in cluttered environments.

\begin{figure}[htbp]
    \centering

    \begin{subfigure}[t]{0.49\linewidth}
        \centering
        \includegraphics[height=4.85cm]{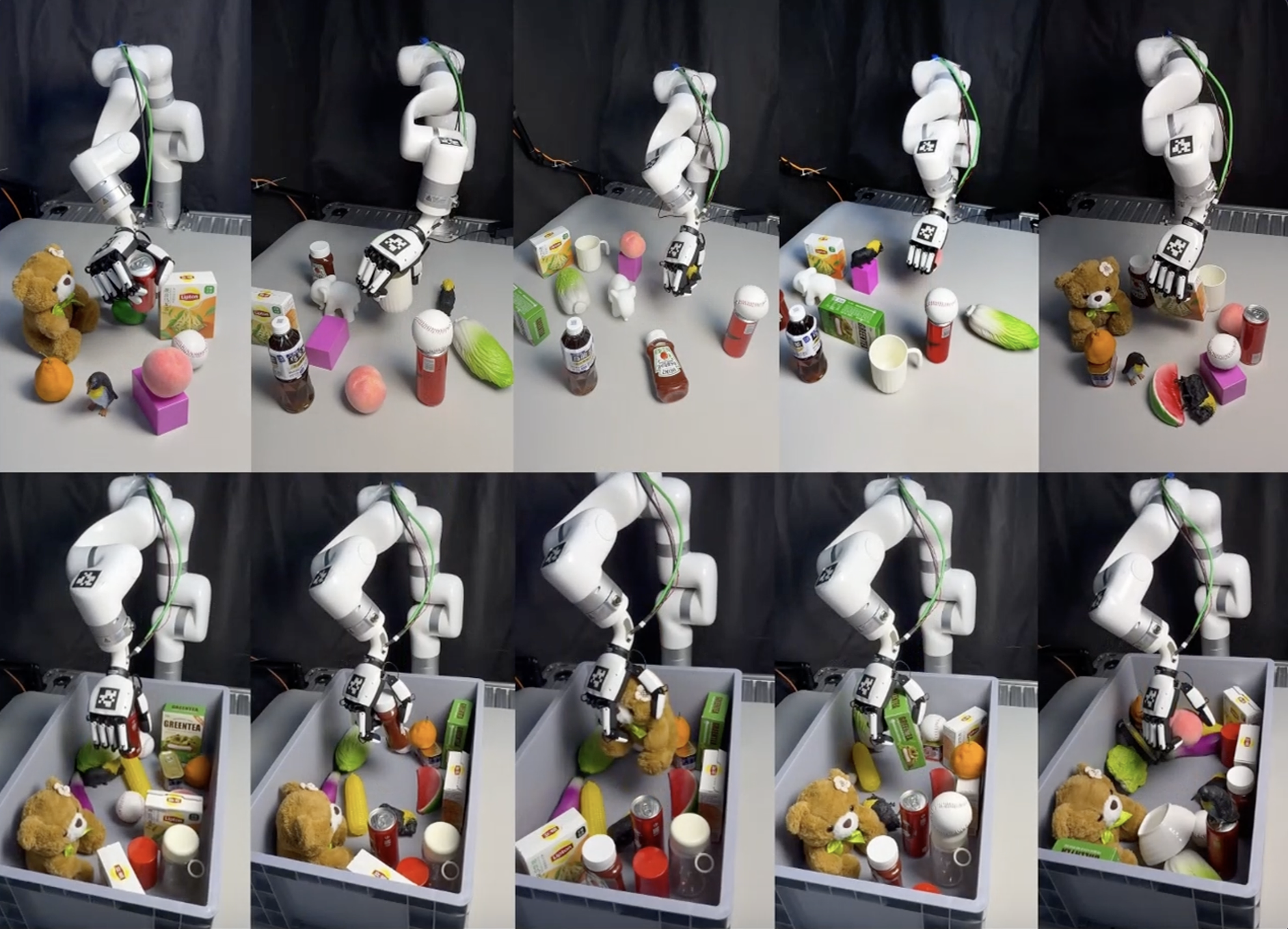}
        \caption{\small XHand}
    \end{subfigure}
    \hfill
    \begin{subfigure}[t]{0.49\linewidth}
        \centering
        \includegraphics[height=4.85cm]{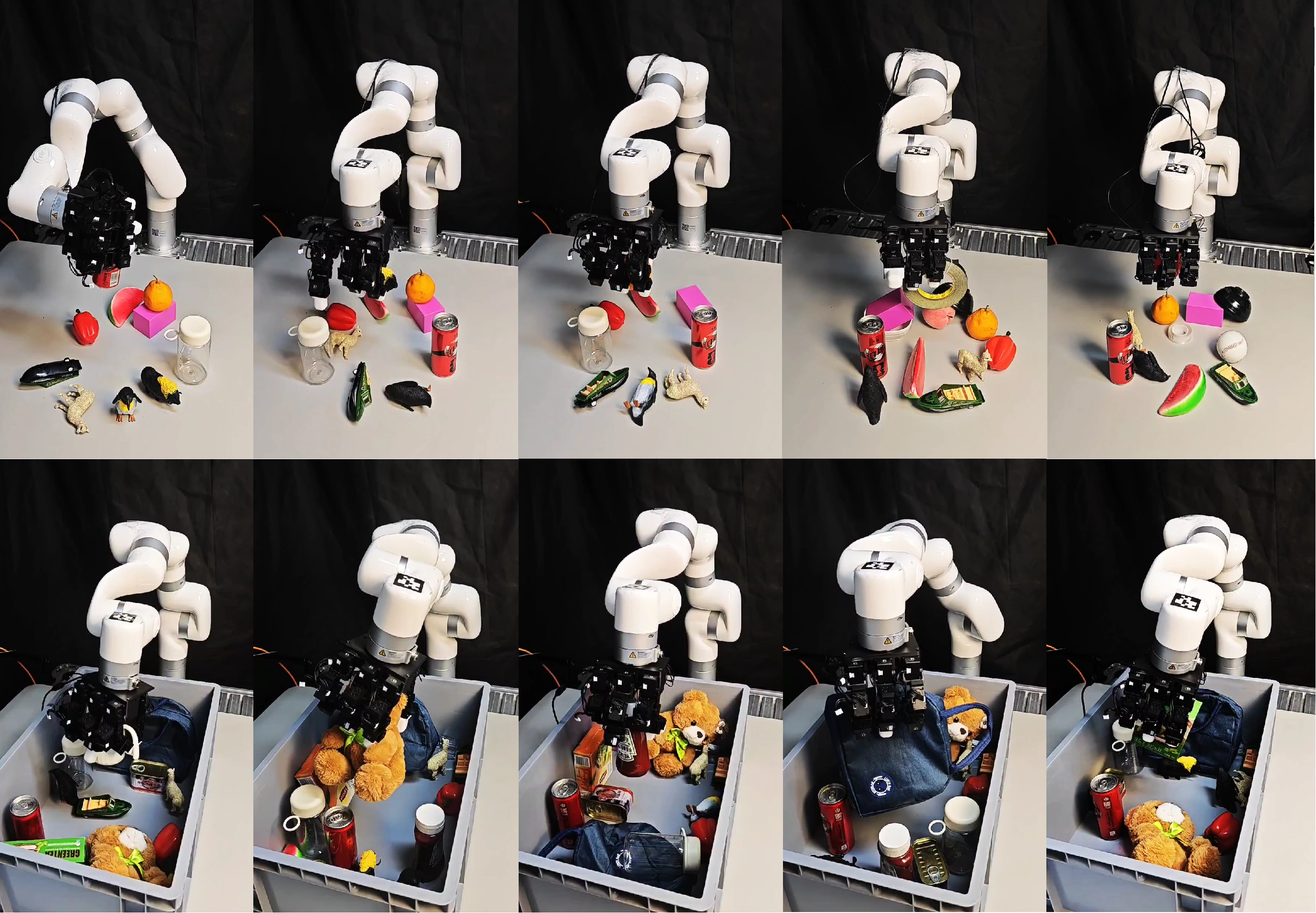}
        \caption{\small LEAP Hand}
    \end{subfigure}

    \caption{\small \textbf{Visualization of real-world experiment results on XHand and LEAP Hand.}}
    \label{fig:realworld_results}
\end{figure}

\paragraph{Experiment Result.}
The real-world results on XHand and LEAP Hand are reported in Tab.~\ref{tab:xhand_results} and Tab.~\ref{tab:leaphand_results}. $\mathcal{T}(R,O)$ Grasp++ achieves overall success rates of 85\% and 82\% on the two hands under cluttered in-box and on-table settings, respectively. As illustrated in Fig.~\ref{fig:realworld_results}, the robot executes stable grasps on diverse daily objects under real-world settings, where observations are often partial and noisy. These experiments demonstrate the robustness of $\mathcal{T}(R,O)$ Grasp++ in challenging real-world manipulation scenes. Video results are provided on the project website~\url{https://nus-lins-lab.github.io/TROGrasp2Web/}.

\begin{table}[htbp]
    \centering
    \label{tab:real_world_results}

    \begin{subtable}[t]{0.49\linewidth}
        \centering
        \scriptsize
        \setlength{\tabcolsep}{1.2pt}
        \renewcommand{\arraystretch}{1.15}

        \begin{tabular}{@{}*{5}{P{0.18\linewidth}}@{}}
            \toprule
            \multicolumn{5}{c}{\textbf{In-Box Setting}} \\
            \midrule
            \obj{Snack}{Box} &
            \obj{Coke}{Bottle} &
            \obj{Peach}{Model} &
            \obj{Ketchup}{\phantom{Box}} &
            \obj{Bear}{Doll} \\
            10/10 & 9/10 & 7/10 & 7/10 & 10/10 \\
            \midrule
            \multicolumn{5}{c}{\textbf{On-Table Setting}} \\
            \midrule
            \obj{Coke}{Bottle} &
            \obj{Peach}{Model} &
            \obj{Animal}{Toy} &
            \obj{Handled}{Cup} &
            \obj{Snack}{Box} \\
            10/10 & 8/10 & 7/10 & 8/10 & 9/10 \\
            \bottomrule
        \end{tabular}
        \caption{\small XHand}
        \label{tab:xhand_results}
    \end{subtable}
    \hfill
    \begin{subtable}[t]{0.49\linewidth}
        \centering
        \scriptsize
        \setlength{\tabcolsep}{1.2pt}
        \renewcommand{\arraystretch}{1.15}

        \begin{tabular}{@{}*{5}{P{0.18\linewidth}}@{}}
            \toprule
            \multicolumn{5}{c}{\textbf{In-Box Setting}} \\
            \midrule
            \obj{Snack}{Box} &
            \obj{Cloth}{Bag} &
            \obj{Handled}{Cup} &
            \obj{Ketchup}{\phantom{Box}} &
            \obj{Bear}{Doll} \\
            9/10 & 10/10 & 7/10 & 7/10 & 9/10 \\
            \midrule
            \multicolumn{5}{c}{\textbf{On-Table Setting}} \\
            \midrule
            \obj{Coke}{Bottle} &
            \obj{Pepper}{\phantom{Box}} &
            \obj{Animal}{Toy} &
            \obj{Tape}{Roll} &
            \obj{Orange}{\phantom{Box}} \\
            8/10 & 7/10 & 7/10 & 10/10 & 9/10 \\
            \bottomrule
        \end{tabular}
        \caption{\small LEAP Hand}
        \label{tab:leaphand_results}
    \end{subtable}
    \caption{\small \textbf{Real-world experiment results on XHand and LEAP Hand.}}
\end{table}

\vspace{-10pt}

\paragraph{Failure Analysis.}

Across more than 40 real-world objects with approximately 10 trials per object, we conduct a detailed failure analysis for Wuji and XHand, as summarized in Table~\ref{tab:failure_mode_analysis}. We categorize failures into four types: \textbf{Prediction} (invalid grasp pose), \textbf{Planning} (motion planning failure), \textbf{Collision} (collision during execution), and \textbf{Instability} (object slippage). Among these failure modes, execution instability is the most prevalent, with nearly half of all failures caused by object slippage despite a valid grasp prediction. This suggests that further improving grasp stability during physical execution is an important direction for enhancing real-world performance.

\begin{table}[htbp]
    \centering
    \fontsize{8.5pt}{10pt}\selectfont
    \renewcommand{\arraystretch}{1.0}
    \setlength{\tabcolsep}{2.0pt}

    \newcommand{\tabrow}{\rule{0pt}{2.4ex}}

    \begin{tabular}{l|cccc|c}
        \toprule
        \tabrow
        \textbf{Hand}
        & \textbf{Prediction}
        & \textbf{Planning}
        & \textbf{Collision}
        & \textbf{Instability}
        & \textbf{Total}
        \\
        \midrule
        \tabrow
        Wuji
        & 13/414
        & 5/414
        & 12/414
        & 44/414
        & 74/414
        \\
        \tabrow
        XHand
        & 22/548
        & 14/548
        & 27/548
        & 29/548
        & 92/548
        \\
        \bottomrule
    \end{tabular}
    \vspace{5pt}
    \caption{\small \textbf{Failure mode analysis in the real world.}}
    \label{tab:failure_mode_analysis}
    
\end{table}

\vspace{-10pt}

\section{Benchmark Details}

\subsection{Data Annotation Pipeline}

As introduced in the main paper, we use a three-stage pipeline to generate dexterous grasp annotations in cluttered scenes. We first train embodiment-specific RL policies to roll out power and precision grasps on floating objects. We then sample stable tabletop object poses and transform each floating-object grasp accordingly, filtering out candidates with tabletop penetration. Finally, we construct tabletop, shelf, and box clutter by placing obstacle objects outside the 2D convex hull of the hand--object grasp configuration. In this section, we illustrate the details of each stage.

\paragraph{Stage 1: Floating Grasp Annotation.}
We first partition the complete object set into small subsets, each containing at most 128 objects. For each dexterous hand and each object subset, we train two PPO~\cite{schulman2017proximal} policies in Isaac Gym to roll out power and precision grasps on floating target objects. The hand is initialized from sampled approaching poses around the object convex hull, and the policy controls both the 6D wrist motion and the actuated hand joints. Each episode consists of a grasping phase followed by a lifting and perturbation phase, where successful grasps are saved as floating-object grasp annotations. The two policies share the same PPO framework but use different reward designs. For the power-grasp policy, the reward encourages the hand to approach the object while regularizing the hand posture, and gives a terminal reward for successful lifting:
{
\setlength{\abovedisplayskip}{8pt}
\setlength{\belowdisplayskip}{8pt}
\begin{equation}
\label{eq:power_reward}
    r_{\text{power}}
    =
    w_q
    \left\|
    \mathbf{q}_t - \mathbf{q}_{\text{prior}}
    \right\|_2
    +
    w_d
    \left(
    \bar{d}_{t-1} - \bar{d}_{t}
    \right)
    (1-\mathbb{I}_{\text{timeout}})
    +
    w_s \mathbb{I}_{\text{succ}},
\end{equation}
}
where $\mathbf{q}_t$ is the current hand joint configuration, $\mathbf{q}_{\text{prior}}$ is the prior hand configuration, $\bar{d}_t$ is the mean distance from hand contact points to the object point cloud, and $\mathbb{I}_{\text{succ}}$ denotes final grasp success. For the precision-grasp policy, we introduce a palm-contact penalty that discourages excessive palm involvement and promotes fingertip contacts:
{
\setlength{\abovedisplayskip}{4pt}
\setlength{\belowdisplayskip}{4pt}
\begin{equation}
\label{eq:precision_reward}
    r_{\text{precision}}
    =
    r_{\text{power}}
    -
    w_p
    \operatorname{clip}
    \left(
    \frac{
    \tau_p - \bar{d}_{\text{palm}}
    }{
    \tau_p
    },
    0,
    1
    \right),
\end{equation}
}
where $\bar{d}_{\text{palm}}$ is the mean distance from palm contact points to the object point cloud, and $\tau_p$ is the palm-contact threshold. Thus, the power-grasp policy favors stable enclosing grasps, while the precision-grasp policy biases toward fingertip contacts. To further improve annotation diversity, we maintain a memory bank of successful grasp features and add a diversity reward $r_{\text{div}}$ that favors grasps different from previous successful samples.

\paragraph{Stage 2: Tabletop Grasp Filtering.}

After obtaining floating-object grasps, we generate stable object poses on a tabletop in simulation. For each target object, we randomly sample 128 initial object poses above the table and let the object fall under gravity until it reaches a stable resting pose. Each successful floating-object grasp is then transformed according to the resulting object pose, so that the relative transformation between the hand and the target object is preserved. We further filter out candidates that cause penetration between tabletop and the dexterous hand. After the second stage, the remaining grasp annotations provide physically plausible and collision-free tabletop grasps for the target object.

\paragraph{Stage 3: Cluttered Scene Construction.}

Finally, we construct cluttered manipulation scenes by adding obstacle objects around each transformed hand-object configuration. For each transformed grasp, we project the target-object point cloud and hand point cloud onto the tabletop plane, and compute their 2D convex hull as a forbidden placement region that approximates the occupied grasp area. Additional obstacle objects are sampled outside this region to introduce surrounding clutter without collision. We consider three scene types: tabletop, shelf, and box. The tabletop scene uses a square support plane with side length in $[0.70,1.00]$ m; the shelf scene has length $[0.80,1.30]$ m, depth $[0.20,0.40]$ m, and height $[0.35,0.50]$ m; and the box scene has side length $[0.70,1.00]$ m and wall height $[0.25,0.35]$ m. For each scene, we randomly place 15 to 25 obstacle objects, simulate them until they settle, and discard invalid obstacles that collide with the forbidden region or fall outside the environment boundary. The final annotation stores the environment point cloud, target-object pose, obstacle poses, and hand joint values.

\subsection{Hierarchical Difficulty Score}

In the proposed $\mathcal{T}(R,O)$ Grasp++ benchmark, we quantify the execution difficulty of each grasp annotation in the cluttered scene from two complementary perspectives: object-level and environment-level difficulty. Object-level difficulty measures the intrinsic grasping challenge posed by the target object, which is determined by its geometry, shape and size. We quantify it as the number of RL optimization steps required for a policy to reach a 50\% success rate on the target object. We define
{
\setlength{\abovedisplayskip}{6pt}
\setlength{\belowdisplayskip}{6pt}
\begin{equation}
\label{eq:object_difficulty}
D_{\text{obj}} =
\operatorname{Norm}
\left(
\min\left(
N_{\text{power}}^{0.5}, N_{\text{precision}}^{0.5}
\right)
\right)
\in (0,1],
\end{equation}
}
where $N_{\text{power}}^{0.5}$ and $N_{\text{precision}}^{0.5}$ denote the numbers of optimization steps required by the power-grasp and precision-grasp policies, respectively, to achieve a 50\% success rate on the target object, and $\operatorname{Norm}(\cdot)$ is normalization operation. If neither policy reaches a 50\% success rate, we set $D_{\text{obj}}=1$.

For environment-level difficulty, we quantify scene clutter using obstacle proximity and spatial enclosure around the target object. Let $\mathbf{P}_{\text{obj}}$ and $\mathbf{P}_{\text{env}}$ denote the point clouds of the target object and the surrounding environment, respectively. We consider environment points within a radius $r_{\max}$ from the target object center, where $r_{\max}$ defines the local neighborhood used for clutter estimation. We first compute the minimum object-environment distance and define the proximity term as a normalized exponential distance score:
{
\setlength{\abovedisplayskip}{3pt}
\setlength{\belowdisplayskip}{3pt}
\begin{equation}
\label{eq:proximity}
d_{\min}
=
\min_{\mathbf{p}\in \mathbf{P}_{\text{obj}},\,\mathbf{q}\in \mathbf{P}_{\text{env}}}
\|\mathbf{p}-\mathbf{q}\|_2,
\quad
D_{\text{env}}^{\text{prox}}
=
\max\left(
0,
\frac{
\exp(-d_{\min}/\sigma)-\exp(-r_{\max}/\sigma)
}{
1-\exp(-r_{\max}/\sigma)
}
\right).
\end{equation}
}

To measure spatial enclosure, we use the target object center as the origin and discretize the local surrounding space into azimuth-elevation bins. The enclosure term $D_{\text{env}}^{\text{enc}}$ is defined as the fraction of bins occupied by nearby environment points. The environment-level difficulty is then defined as:
{
\setlength{\abovedisplayskip}{3pt}
\setlength{\belowdisplayskip}{3pt}
\begin{equation}
\label{eq:env_difficulty}
D_{\text{env}}
=
\operatorname{Norm}
\left(
\alpha D_{\text{env}}^{\text{prox}}
+
\beta D_{\text{env}}^{\text{enc}}
\right),
\quad
\alpha+\beta=1.
\end{equation}
}

\begin{figure}[htbp]
    \centering
    \includegraphics[width=\linewidth]{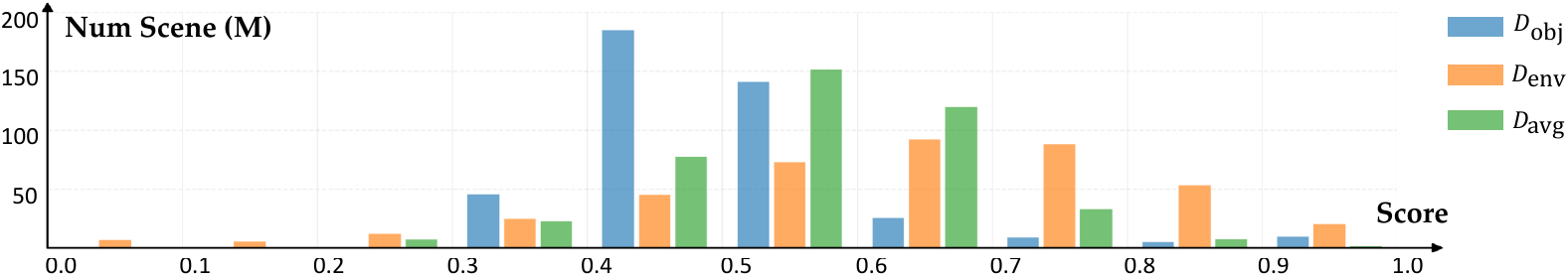}
    \vspace{-15pt}
    \caption{\small \textbf{Score distribution of our proposed benchmark.} Each grasp in a cluttered scene is annotated with a difficulty score derived from object geometry, obstacle proximity, and spatial enclosure.}
    \label{fig:benchmark}
\end{figure}

Combining object-level difficulty $D_{\text{obj}}$ and environment-level difficulty $D_{\text{env}}$, we assign each cluttered-scene grasp an overall difficulty score
$D_{\text{avg}} = (D_{\text{obj}} + D_{\text{env}}) / 2$.
The resulting score distributions over all grasp samples are shown in Fig.~\ref{fig:benchmark}, demonstrating that our benchmark covers a broad range of grasping difficulty at both the object and environment levels. These hierarchical difficulty annotations enable structured evaluation across various clutter levels and object grasping difficulties, helping reveal the performance boundaries of different methods.

\subsection{Dataset Diversity}

To validate the diversity of our proposed benchmark, we sample around 10 million grasps across eight hands and quantify diversity in palm translation/rotation, normalized joint configurations, and approach directions, as shown in Table~\ref{tab:diversity_analysis}. Approach diversity is measured by entropy over six Cartesian directions ($\pm x,\pm y,\pm z$), scaled to $[0,6]$. The results demonstrate that our large-scale RL pipeline generates highly diverse grasps across embodiments, configurations, and approach directions, rather than collapsing to a few dominant modes.

\begin{table}[htbp]
    \centering
    \fontsize{8.5pt}{10pt}\selectfont
    \renewcommand{\arraystretch}{1.0}
    \setlength{\tabcolsep}{2.0pt}

    \newcommand{\tabrow}{\rule{0pt}{2.4ex}}

    \begin{tabular}{cccc}
        \toprule
        \tabrow
        \textbf{Translation Div.}
        & \textbf{Rotation Div.}
        & \textbf{Norm. Joint Div.}
        & \textbf{Direction Ent.}
        \\
        \midrule
        \tabrow
        0.360 m
        & 125.97 deg
        & 5.97
        & 5.96 / 6
        \\
        \bottomrule
    \end{tabular}
    \vspace{5pt}
    \caption{\small \textbf{Diversity analysis of $\mathcal{T}(R,O)$++ dataset.}}
    \label{tab:diversity_analysis}
\end{table}
\vspace{-15pt}

\section{Implementation Details}

\paragraph{Diffusion Model Architecture.}

Our diffusion model operates on $\mathcal{T}(R,O)$ Graph++ representation, which is composed of object nodes, environment nodes and robot-link nodes. Object and environment geometries are first encoded into local geometry tokens by pretrained VQ-VAE~\cite{van2017neural} encoders. Across all model scales, we use $P=25$ object patches, $E=25$ environment patches, and at most $L=25$ robot-link nodes. The denoiser takes noisy robot-link poses and graph edge features as input, and predicts link-wise pose updates through stacked graph attention layers. Tab.~\ref{tab:diffusion_model_config} summarizes the detailed network configurations for different model scales. In our experiments, we mainly use the 0.2B model as the default configuration.

\begin{table}[htbp]
    \centering
    \resizebox{0.9\linewidth}{!}{
    \begin{tabular}{cccccc}
        \toprule
        \textbf{Model Params.} & \textbf{Layers Num.} & \textbf{Attn. Heads Num.} & \textbf{Node Dim.} & \textbf{Edge Dim.} & \textbf{Output Layer Dim.} \\
        \midrule
        10M  & 4 & 4 & 256 & 256 & [128, 64] \\
        50M  & 8 & 6 & 384 & 384 & [256, 128] \\
        \textbf{0.2B} & \textbf{8} & \textbf{12} & \textbf{768} & \textbf{768} & \textbf{[512, 256, 128]} \\
        1B   & 12 & 24 & 1536 & 1536 & [512, 256, 128] \\
        \bottomrule
    \end{tabular}
    }
    \vspace{5pt}
    \caption{\small \textbf{Architecture configurations of diffusion models with different parameter scales.}}
    \label{tab:diffusion_model_config}
\end{table}
\vspace{-15pt}

\paragraph{Training Configurations.}

$\mathcal{T}(R,O)$ Grasp++ takes three inputs: the target-object point cloud $\mathbf{P}_{\text{obj}} \in \mathbb{R}^{512 \times 3}$, the surrounding-environment point cloud $\mathbf{P}_{\text{env}} \in \mathbb{R}^{2048 \times 3}$, and the URDF description of an arbitrary dexterous hand. For the default 0.2B configuration, we train $\mathcal{T}(R,O)$ Grasp++ with a batch size of 48 per GPU and set the number of optimization steps to ensure full coverage of the training dataset. We use the AdamW~\cite{loshchilov2017decoupled} optimizer with a learning rate of $1\times10^{-4}$, 1,000 warmup steps, gradient clipping of 1.0, and bfloat16 mixed precision. The diffusion process follows a cosine noise schedule with $T=1,000$ timesteps and offset $s=0.008$. For each training batch, we sample $N_t=5$ diffusion timesteps for supervision. For the loss function, we set loss weights $\gamma_{\text{pos}}=\gamma_{\text{rot}}=1.0$. We train the model on 8 NVIDIA A100 80GB GPUs.

\paragraph{Inference Configurations.}

During inference, we use DDIM sampling~\cite{song2020denoising} with 20 denoising steps and set $\eta=1.0$. Given the point clouds of the target object and surrounding environment, together with the URDF description of the specified dexterous hand, $\mathcal{T}(R,O)$ Grasp++ iteratively denoises noisy robot-link poses and outputs link-wise $\mathrm{SE}(3)$ grasp poses. The predicted link poses are then retargeted to executable joint values based on Pyroki~\cite{kim2025pyroki} toolkit. To improve the accuracy of the retargeting step, we solve IK from 8 different initializations in parallel and select the solution with the lowest average target-link position error after forward-kinematics evaluation.

\section{Additional Experiment}

\subsection{Cross-Embodiment Zero-Shot Generalization}

Although $\mathcal{T}(R,O)$ Grasp++ is trained on eight dexterous embodiments covering 3-finger, 4-finger, and 5-finger hands, the diversity of available dexterous hands remains limited. Therefore, directly evaluating zero-shot generalization on entirely novel hands is challenging. To assess the cross-embodiment zero-shot generalization ability of $\mathcal{T}(R,O)$ Grasp++, we derive new hand embodiments by modifying the link lengths and joint limits of the existing training hands. We define link alignment $S_{\text{link}}$ and joint overlap $S_{\text{joint}}$ to measure the similarity between each derived hand and its original dexterous embodiment as follows:
{
\setlength{\abovedisplayskip}{5pt}
\setlength{\belowdisplayskip}{5pt}
\begin{equation}
S_{\text{link}} = 1 - \frac{1}{L}\sum\nolimits_{i=1}^L\frac{|l_i'-l_i|}{l_i}, \quad
S_{\text{joint}} = \frac{1}{L}\sum\nolimits_{i=1}^L \frac{|j_i'\cap j_i|}{|j_i'\cup j_i|}.
\label{eq:embodiment similarity}
\end{equation}
}

where $l_i$ and $l_i'$ denote the original and modified link lengths, $j_i$ and $j_i'$ are the corresponding joint ranges. Under this definition, we conduct cross-embodiment zero-shot experiment on XHand. 

\begin{table}[htbp]
    \centering
    \resizebox{0.6\linewidth}{!}{
    \begin{tabular}{ccccc}
        \toprule
        \multirow{2}{*}{\textbf{Similarity}} 
        & \multicolumn{2}{c}{\textbf{Link Alignment} $S_{\text{link}}$} 
        & \multicolumn{2}{c}{\textbf{Joint Overlap} $S_{\text{joint}}$} \\
        \cmidrule(lr){2-3} \cmidrule(lr){4-5}
        & \textbf{SR (\%) $\uparrow$} & \textbf{Diversity $\uparrow$}
        & \textbf{SR (\%) $\uparrow$} & \textbf{Diversity $\uparrow$} \\
        \midrule
        1.0 & 86.03 & 0.418 & 86.03 & 0.418 \\
        0.9 & 85.35 & 0.413 & 84.04 & 0.411 \\
        0.8 & 84.56 & 0.411 & 82.93 & 0.400 \\
        0.7 & 84.51 & 0.412 & 81.73 & 0.390 \\
        \bottomrule
    \end{tabular}
    }
    \vspace{4pt}
    \caption{\small \textbf{Cross-embodiment zero-shot performance.}}
    \label{tab:zero_shot}
\end{table}

As shown in Tab.~\ref{tab:zero_shot}, $\mathcal{T}(R,O)$ Grasp++ maintains strong zero-shot generalization to unseen hand variants. When the embodiment similarity decreases from 1.0 to 0.7, the success rate remains above 80\% for both link length and joint range perturbations. These results indicate that $\mathcal{T}(R,O)$ Grasp++ can generalize to novel dexterous embodiments with modified kinematic structures, demonstrating its effectiveness as a unified cross-embodiment grasping model.

\end{document}